%% file: eccv2020submissionCR.tex
\documentclass[runningheads]{llncs}
\usepackage{graphicx}
\usepackage{comment}
\usepackage{amsmath,amssymb} 
\usepackage{color}

\usepackage{wrapfig}

\input{macros.tex}

\begin{document}
\pagestyle{headings}
\mainmatter
\def\ECCVSubNumber{2918}  

\title{Contact and Human Dynamics from \\ Monocular Video} 

\titlerunning{Contact and Human Dynamics from Monocular Video}
%
\author{Davis Rempe\inst{1,2} \and
Leonidas J. Guibas\inst{1} \and
Aaron Hertzmann\inst{2} \and
Bryan Russell\inst{2} \and
Ruben Villegas\inst{2} \and
Jimei Yang\inst{2}}
\index{Guibas, Leonidas}
\authorrunning{D. Rempe et al.}
%
\institute{Stanford University \and
Adobe Research \\
\email{\href{https://geometry.stanford.edu/projects/human-dynamics-eccv-2020/}{geometry.stanford.edu/projects/human-dynamics-eccv-2020}}
}
\maketitle

\input{content/text/00_abstract.tex}
\input{content/text/01_introduction.tex}
\input{content/text/02_related_work.tex}
\input{content/text/03_motion_estimation.tex}
\input{content/text/04_experiments.tex}
\input{content/text/05_discussion.tex}

\clearpage

\acknowledgments{} This work was in part supported by NSF grant IIS-1763268, grants from the Samsung GRO program and the Stanford SAIL Toyota Research Center, and a gift from Adobe Corporation. We thank the following YouTube channels for sharing their videos online: Dance FreaX, Dancercise Studio, Fencer’s Edge, MihranTV, DANCE TUTORIALS, Deepak Tulsyan, Gibson Moraes, and pigmie.

{\small
\bibliographystyle{splncs04}
\bibliography{egbib}
}

\clearpage

\input{appendix/macros}

\section*{Appendices}

\input{appendix/text/02_contact.tex}
\input{appendix/text/03_kinematic.tex}
\input{appendix/text/05_physics_optim.tex}
\input{appendix/text/04_retargeting.tex}
\input{appendix/text/07_results.tex}

\end{document}

%% file: macros.tex
\usepackage[dvipsnames]{xcolor}
\usepackage{amsfonts}
\usepackage{bm}
\usepackage{amsmath}
\usepackage{multirow}
\usepackage{booktabs}
\usepackage{hyperref}

\newcommand{\ie}{i.e.}
\newcommand{\etal}{et al.}
\newcommand{\eg}{e.g.}

\newcommand{\acknowledgments}[1]{\noindent{\textbf{Acknowledgments. }}}

\newcommand{\E}[1]{E_{\mathit{#1}}}

\newcommand{\bg}{\mathbf{g}}
\newcommand{\bx}{\mathbf{x}}

\newcommand{\bS}{\mathbf{S}}

\newcommand{\bp}{\mathbf{p}}

\newcommand{\bI}{\mathbf{I}}
\newcommand{\bq}{\mathbf{q}}
\newcommand{\br}{\mathbf{r}}
\newcommand{\bn}{\mathbf{n}}
\newcommand{\bt}{\mathbf{t}}
\newcommand{\beff}{\mathbf{f}}

\newcommand{\btheta}{\bm{\theta}}
\newcommand{\reals}{\mathbb{R}}
\newcommand{\bomega}{\bm{\omega}}

\newcommand{\brb}{\overline{\mathbf{r}}}

\newcommand{\bpb}{\overline{\mathbf{p}}}
\newcommand{\bthetab}{\overline{\bm{\theta}}}
\newcommand{\mindent}{\bm{\;\;\;\;\;\;}}

%% file: content/text/00_abstract.tex
\begin{abstract}
Existing deep models predict 2D and 3D kinematic poses from video that are approximately accurate, but contain visible errors that violate physical constraints, such as feet penetrating the ground and bodies leaning at extreme angles. In this paper, we present a physics-based method for inferring 3D human motion from video sequences that takes initial 2D and 3D pose estimates as input. We first estimate ground contact timings with a novel prediction network which is trained without hand-labeled data. A physics-based trajectory optimization then solves for a physically-plausible motion, based on the inputs. We show this process produces motions that are significantly more realistic than those from purely kinematic methods, substantially improving quantitative measures of both kinematic and dynamic plausibility. We demonstrate our method on character animation and pose estimation tasks on dynamic motions of dancing and sports with complex contact patterns.
\end{abstract}

%% file: content/text/01_introduction.tex
\section{Introduction}
Recent methods for human pose estimation from monocular video \cite{Bogo:ECCV:2016,hmrKanazawa17,SMPL-X:2019,mtc} estimate accurate overall body pose with small absolute differences from the true poses in body-frame 3D coordinates. However, the recovered motions in world-frame are visually and physically implausible in many ways, including feet that float slightly or penetrate the ground,  implausible forward or backward body lean, and motion errors like jittery, vibrating poses.
These errors would prevent many subsequent uses of the motions. For example, inference of actions, intentions, and emotion often depends on subtleties of pose, contact and acceleration, 
as does computer animation; human perception is highly sensitive to physical inaccuracies \cite{Hoyet:2012:PRP,Reitsma:2003:PMC}.  Adding more training data would not solve these problems, because existing methods do not account for physical plausibility.

Physics-based trajectory optimization presents an appealing solution to these issues, particularly for dynamic motions like walking or dancing. Physics imposes important constraints that are hard to express in pose space but easy in terms of dynamics. For example, feet in static contact do not move, the body moves smoothly overall relative to contacts, and joint torques are not large. However, full-body dynamics is notoriously difficult to optimize~\cite{Safonova:2004}, in part because contact is discontinuous, and the number of possible contact events grows exponentially in time. As a result, combined optimization of contact and dynamics is enormously sensitive to local minima. 

\input{content/figures/teaser.tex}

This paper introduces a new strategy for extracting dynamically valid full-body motions from monocular video (Figure \ref{fig:teaser}), combining learned pose estimation with physical reasoning through trajectory optimization. As input, we use the results of kinematic pose estimation techniques \cite{cao2018openpose,mtc}, which produce accurate overall poses but inaccurate contacts and dynamics. Our method leverages a reduced-dimensional body model with centroidal dynamics and contact constraints \cite{dai2014whole,winkler18} to produce a physically-valid motion that closely matches these inputs. We first infer foot contacts from 2D poses in the input video which are then used in a physics-based trajectory optimization to estimate 6D center-of-mass motion, feet positions, and contact forces. We show that a contact prediction network can be accurately trained on synthetic data. This allows us to separate initial contact estimation from motion optimization, making the optimization more tractable. As a result, our method is able to handle highly dynamic motions without sacrificing physical accuracy.

We focus on single-person dynamic motions from dance, walking, and sports. Our approach substantially improves the realism of inferred motions over state-of-the-art methods, and estimates numerous physical properties that could be useful for further inference of scene properties and action recognition. We primarily demonstrate our method on character animation by retargeting captured motion from video to a virtual character. We evaluate our approach using numerous kinematics and dynamics metrics designed to measure the physical plausiblity of the estimated motion. The proposed method takes an important step to incorporating physical constraints into human motion estimation from video, and shows the potential to reconstruct realistic, dynamic sequences.

%% file: content/figures/teaser.tex
\begin{figure}[t]
\begin{center}
\includegraphics[width=0.8\linewidth]{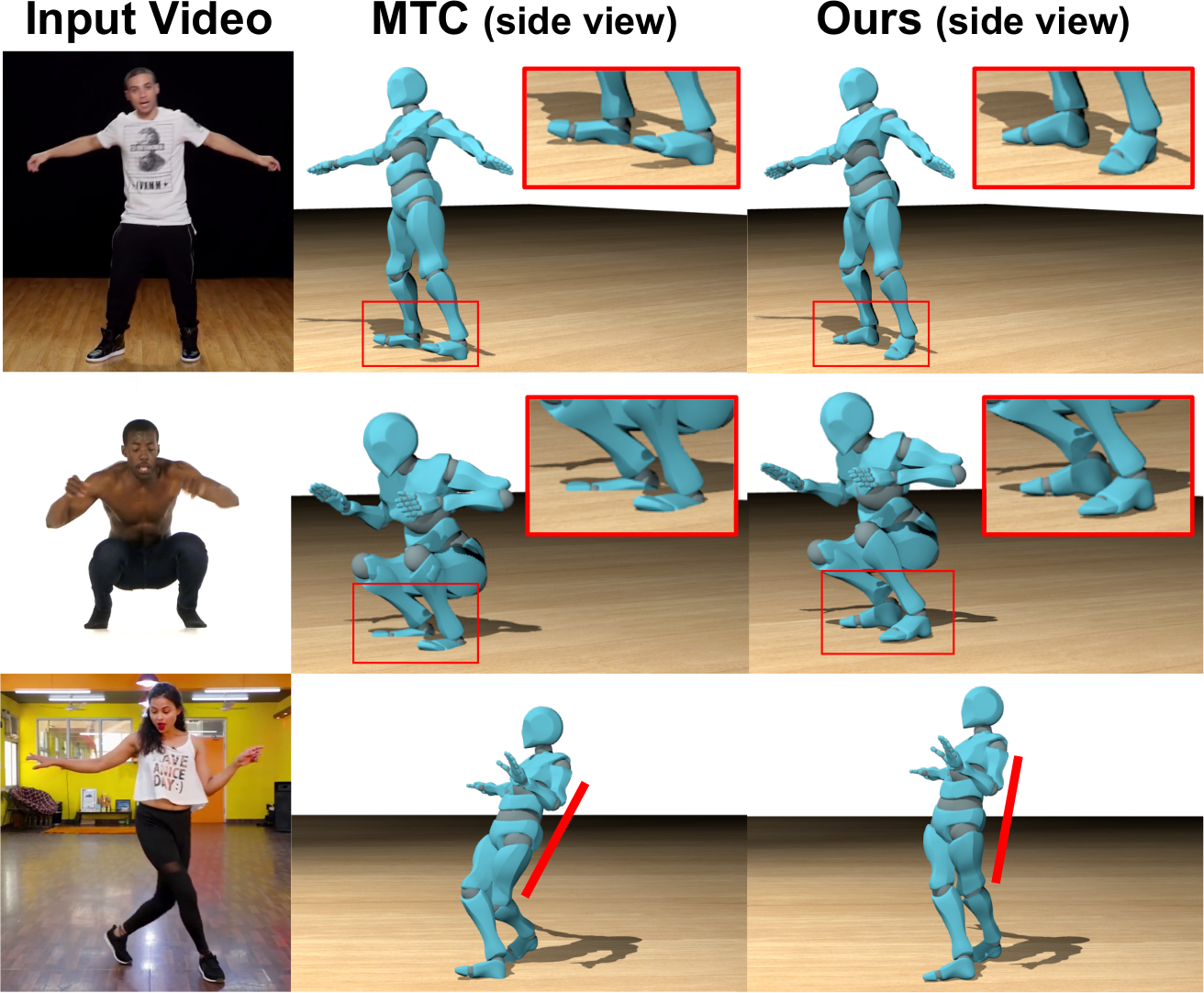}
\end{center}
   \caption{Our contact prediction and physics-based optimization corrects numerous physically implausible artifacts common in 3D human motion estimations from, \eg, Monocular Total Capture (MTC) \cite{mtc} such as foot floating (top row), foot penetrations (middle), and unnatural leaning (bottom).
   }
\label{fig:teaser}
\end{figure}

%% file: content/text/02_related_work.tex
\section{Related Work}
We build on several threads of work in computer vision, animation, and robotics, each with a long history \cite{ForsythSurvey}. Recent vision results are detailed here.

Recent progress in pose estimation can accurately detect 2D human keypoints~\cite{cao2018openpose,he2017maskrcnn,newell2016stacked} and infer 3D pose~\cite{Bogo:ECCV:2016,hmrKanazawa17,SMPL-X:2019} from a single image. Several recent methods extract 3D human motions from monocular videos by exploring various forms of temporal cues~\cite{humanMotionKanazawa19,VNect_SIGGRAPH2017,Xu:2018:MHP,mtc}.
While these methods focus on explaining human motion in pixel space, they do not account for physical plausibility. Several recent works interpret interactions between people and their environment in order to make inferences about each \cite{chen2019holistic++,hassan2019resolving,zanfir2018monocular}; each of these works uses only static kinematic constraints. Zou \etal~\cite{Zou_2020_WACV} infer contact constraints to optimize 3D motion from video. We show how dynamics can improve inference of human-scene interactions, leading to more physically plausible motion capture.

Some works have proposed physics constraints to address the issues of kinematic tracking. Brubaker \etal~\cite{BrubakerIJCV2010} propose a physics-based tracker based on a reduced-dimensional walking model. Wei and Chai \cite{Wei:2010:VMP} track body motion from video, assuming keyframe and contact constraints are provided.  Similar to our own work, Brubaker and Fleet \cite{BrubakerICCV2009} perform trajectory optimization for full-body motion. To jointly optimize contact and dynamics, they use a continuous approximation to contact. However, soft contact models introduce new difficulties, including inaccurate transitions and sensitivity to stiffness parameters, while still suffering from local minima issues. Moreover, their reduced-dimensional model includes only center-of-mass positional motion, which does not handle rotational motion well. In contrast, we obtain accurate contact initialization in a preprocessing step to simplify optimization, and we model rotational inertia. 

Li \etal\ \cite{Li_2019_CVPR} estimate dynamic properties from videos. We share the same overall pipeline of estimating pose and contacts, followed by trajectory optimization. Whereas they focus on the dynamics of human-object interactions, we focus on videos where the human motion itself is much more dynamic, with complex variation in pose and foot contact; we do not consider human-object interaction. They use a simpler data term, and perform trajectory optimization in full-body dynamics unlike our reduced representation. Their classifier training requires hand-labeled data, unlike our automatic dataset creation method.

Prior methods learn character animation controllers from video. Vondrak \etal\ \cite{Vondrak:2012:VMC} train a state-machine controller using image silhouette features.
Peng \etal\ \cite{Peng:2018:SRL} train a controller to perform skills by following kinematically-estimated poses from input video sequences. They demonstrate impressive results on a variety of skills. They do not attempt accurate reconstruction of motion or contact, nor do they evaluate for these tasks, rather they focus on control learning.

Our optimization is related to physics-based methods in  computer animation, e.g., \cite{Fang:2003:ESP,Jiang:2019:SBR,deLasa:2010:FLC,Liu:2005:LPM,Macchietto:2009:MCB,Popovic:1999:PBM,biolocomotion}. Two unique features of our optimization are the use of low-dimensional dynamics optimization that includes 6D center-of-mass motion and contact constraints, thereby capturing important rotational and footstep quantities without requiring full-body optimization, and the use of a classifier to determine contacts before optimization.

%% file: content/text/03_motion_estimation.tex
\section{Physics-Based Motion Estimation}
This section describes our approach, which is 
summarized in Figure~\ref{fig:flowchart}.
The core of our method is a physics-based trajectory optimization that enforces dynamics on the input motion (Section \ref{section:physicsoptim}).  Foot contact timings are estimated in a preprocess (Section \ref{section:contact}), along with other inputs to the optimization (Section \ref{sec:kinematic}). Similar to previous work \cite{Li_2019_CVPR,mtc}, in order to recover full-body motion we assume there is no camera motion and that the full body is visible.

\input{content/figures/pipeline.tex}

\subsection{Physics-Based Trajectory Optimization}\label{section:physicsoptim}

The core of our framework is an optimization which enforces dynamics on an initial motion estimate given as input (see Section~\ref{sec:kinematic}). 
The goal is to improve the plausibility of the motion by applying physical reasoning through the objective and constraints. 
We aim to avoid common perceptual errors, \eg, jittery, unnatural motion with feet skating and ground penetration, by generating a smooth trajectory with physically-valid momentum and static feet during contact.

The optimization is performed on a reduced-dimensional body model that captures overall motion, rotation, and contacts, but avoids the difficulty of optimizing all joints. Modeling rotation is necessary for important effects like arm swing and counter-oscillations \cite{HerrPopovic,deLasa:2010:FLC,Macchietto:2009:MCB}, and the reduced-dimensional \emph{centroidal} dynamics model can produce plausible trajectories for humanoid robots~\cite{carpentier2018multicontact,dai2014whole,orin2013centroidal}.
Our method is based on a recent robot motion planning algorithm from Winkler \etal\ \cite{winkler18} that leverages a simplified version of centroidal dynamics, which treats the robot as a rigid body with a fixed mass and moment of inertia. Their method finds a feasible trajectory by optimizing the position and rotation of the center-of-mass (COM) along with feet positions, contact forces, and contact durations as described in detail below. 
We modify this algorithm to suit our computer vision task: we use a temporally varying inertia tensor which allows for changes in mass distribution (swinging arms) and enables estimating the dynamic motions of interest, we add energy terms to match the input kinematic motion and foot contacts, and we add new kinematics constraints for our humanoid skeleton. 
\paragraph{Inputs.}
The method takes initial estimates of: COM position $\bar{\br}(t) \in \reals^3$ and orientation $\bar{\btheta}(t) \in \reals^3$ trajectories, body-frame inertia tensor trajectory $\bI_{b}(t) \in \reals^{3\times3}$, and trajectories of the foot joint positions $\bar{\bp}_{1:4}(t) \in \reals^3$. 
There are four foot joints: left toe base, left heel, right toe base, and right heel, indexed as $i\in\{1,2,3,4\}$. These inputs are at discrete timesteps, but we write them here as functions for clarity.
The 3D ground plane height $h_{\mathit{floor}}$ and upward normal is provided.
Additionally, for each foot joint at each time, a binary label is provided indicating whether the foot is in contact with the ground. These labels determine initial estimates of contact durations for each foot joint $\bar{T}_{i, 1}, \bar{T}_{i, 2}, \dots, \bar{T}_{i, n_{i}}$ as described below.
The distance from toe to heel $\ell_{\mathit{foot}}$ and maximum distance from toe to hip $\ell_{\mathit{leg}}$ are also provided.
All quantities are computed from video input as described in Sections \ref{section:contact} and \ref{sec:kinematic}, and are used to both initialize the optimization variables and as targets in the objective function.
\paragraph{Optimization Variables.}
The optimization variables are the COM position and Euler angle orientation $\br(t), \btheta(t) \in \reals^3$, foot joint positions $\bp_i(t) \in \reals^3$ and contact forces $\beff_i(t) \in \reals^3$.
These variables are continuous functions of time, represented by piece-wise cubic polynomials with continuity constraints. We also optimize contact timings. The contacts for each foot joint are independently parameterized by a sequence of phases that alternate between contact and flight. The optimizer cannot change the type of each phase (contact or flight), but it can modify their durations $T_{i, 1}, T_{i, 2}, \dots, T_{i, n_{i}} \in \reals$ where $n_{i}$ is the number of total contact phases for the $i$th foot joint.
\paragraph{Objective.}
Our complete formulation is shown in Figure \ref{fig:physicsformulation}. $E_{\mathit{data}}$ and $E_{\mathit{dur}}$ seek to keep the motion and contacts as close as possible to the intial inputs, which are derived from video, at discrete steps over the entire duration $T$:
\begin{align}
    E_{\mathit{data}}(t) &= w_r || \br(t) -  \brb(t) ||^2 + w_\theta || \btheta(t) - \bthetab(t) ||^2 \nonumber \\ 
    &+ w_p \sum_{i=1}^{4} || \bp_i(t) - \bpb_i(t) ||^2 \\
    E_{\mathit{dur}} &= w_d \sum_{i=1}^{4} \sum_{j=1}^{n_i} (T_{i,j} - \bar{T}_{i,j})^2
\end{align}
We weigh these terms with $w_d = 0.1$, $w_r = 0.4$, $w_\theta = 1.7$, $w_p = 0.3$.

The remaining objective terms are regularizers that prefer small velocities and accelerations resulting in a smoother optimal trajectory:
\begin{align}
    E_{\mathit{vel}}(t) &= \gamma_{r}|| \dot{\br}(t) ||^2 + \gamma_{\theta}|| \dot{\btheta}(t) ||^2 + \gamma_p \sum_{i=1}^{4} || \dot{\bp}_i(t) ||^2 \\
    E_{\mathit{acc}}(t) &= \beta_{r} || \ddot{\br}(t) ||^2 + \beta_{\theta}|| \ddot{\btheta}(t) ||^2 + \beta_{p} \sum_{i=1}^{4} || \ddot{\bp}_i(t) ||^2
\end{align}
with $\gamma_{r} = \gamma_{\theta} = 10^{-3}$, $\gamma_{p} = 0.1$ and $\beta_{r} = \beta_{\theta} = \beta_{p} = 10^{-4}$.

\input{content/figures/physics_formulation.tex}

\paragraph{Constraints.}
The first set of constraints strictly enforce valid rigid body mechanics, including linear and angular momentum. 
This enforces important properties of motion, for example, during flight the COM must follow a parabolic arc according to Newton's Second Law. During contact, the body motion acceleration is limited by the possible contact forces \eg, one cannot walk at a 45$^{\circ}$ lean.

At each timestep, we use the world-frame inertia tensor $\bI_w(t)$ computed from the input $\bI_b(t)$ and the current orientation $\btheta(t)$. This assumes that the final output poses will not be dramatically different from those of the input: a reasonable assumption since our optimization does not operate on upper-body joints and changes in feet positioning are typically small (though perceptually important).
We found that using a constant inertia tensor (as in Winkler\ \etal~\cite{winkler18}) made convergence difficult to achieve. The gravity vector is $\bg = -9.8\hat{\bn}$, where $\hat{\bn}$ is the ground normal.  The angular velocity $\bomega$ is a function of the rotations $\btheta$~\cite{winkler18}.

The contact forces are constrained to ensure that they push away from the floor but are not greater than $f_{\mathit{max}} = 1000$ N in the normal direction. 
With 4 feet joints, this allows 4000 N of normal contact force: about the magnitude that a 100 kg (220 lb) person would produce for extremely dynamic dancing motion~\cite{KuligGRF}.
We assume no feet slipping during contact, so forces must also remain in a friction pyramid defined by friction coefficient $\mu = 0.5$ and floor plane tangents $\hat{\bt}_1, \hat{\bt}_2$. Lastly, forces should be zero at any foot joint not in contact.

Foot contact is enforced through constraints. When a foot joint is in contact, it should be stationary (no-slip) and at floor height $h_{\mathit{floor}}$.
When not in contact, feet should always be on or above the ground. This avoids feet skating and penetration with the ground.

In order to make the optimized motion valid for a humanoid skeleton, the toe and heel of each foot should maintain a constant distance of $\ell_{\mathit{foot}}$.
Finally, no foot joint should be farther from its corresponding hip than the length of the leg $\ell_{\mathit{leg}}$. The hip position $\bp_{\mathit{hip,} i}(t)$ is computed from the COM orientation at that time based on the hip offset in the skeleton detailed in Section \ref{sec:kinematic}.
\paragraph{Optimization Algorithm.}
We optimize with IPOPT \cite{ipopt}, a nonlinear interior point optimizer, using analytical derivatives. We perform the optimization in stages: we first use fixed contact phases and no dynamics constraints to fit the polynomial representation for COM and feet position variables as close as possible to the input motion. Next, we add in dynamics constraints to find a physically valid motion, and finally we allow contact phase durations to be optimized to further refine the motion if possible.

Following the optimization, we compute a full-body motion from the physically-valid COM and foot joint positions using Inverse Kinematics (IK) on a desired skeleton $\bS_{\textit{tgt}}$ (see supplementary Appendix~\ref{appendix:physoptim}).

\subsection{Learning to Estimate Contacts}\label{section:contact}
Before performing our physics-based optimization, we need to infer when the subject's feet are in contact with the ground, given an input video. These contacts are a target for the physics optimization objective and their accuracy is crucial to its success. To do so, we train a network that, for each video frame, classifies whether the toe and heel of each foot are in contact with the ground. 

The main challenge is to construct a suitable dataset and feature representation. There is currently no publicly-available dataset of videos with labeled foot contacts and a wide variety of dynamic motions. Manually labeling a large, varied dataset would be difficult and costly. Instead, we generate synthetic data using motion capture (mocap) sequences. We automatically label contacts in the mocap and then use 2D joint position features from OpenPose \cite{cao2018openpose} as input to our model, rather than image features from the raw rendered video frames.  This allows us to train on synthetic data but then apply the model to real inputs. 
\paragraph{Dataset.}
To construct our dataset, we obtained 65 mocap sequences for the 13 most human-like characters from \url{www.mixamo.com}, ranging from dynamic dancing motions to idling.
Our set contains a diverse range of mocap sequences, retargeted to a variety of animated characters. At each time of each motion sequence, four possible contacts are automatically labeled by a heuristic: a toe or heel joint is considered to be in contact when (i) it has moved less than 2 cm from the previous time, and (ii) it is within 5 cm from the known ground plane.
Although more sophisticated labeling \cite{Ikemoto:2006:KPY,LeCallennec:2006:RKC}  could be used, we found this approach sufficiently accurate to learn a model for the videos we evaluated on.

We render these motions (see Figure~\ref{fig:motionresults}(c)) on their rigged characters with motion blur, randomized camera viewpoint, lighting, and floor texture. For each sequence, we render two views, resulting in over 100k frames of video with labeled contacts and 2D and 3D poses.
Finally, we run a 2D pose estimation algorithm, OpenPose \cite{cao2018openpose}, to obtain the 2D skeleton which our model uses as input.
\paragraph{Model and Training.}
The classification problem is to map from 2D pose in each frame to the four contact labels for the feet joints.  As we demonstrate in Section \ref{sec:contactres}, simple heuristics based on 2D velocity do not accurately label contacts due to the ambiguities of 3D projection and noise.

For a given time $t$, our labeling neural network takes as input the 2D poses over a temporal window of duration $w$ centered on the target frame at $t$. The 2D joint positions over the window are normalized to place the root position of the target frame at $(0,0)$, resulting in relative position and velocity.
We set $w=9$ video frames and use the 13 lower-body joint positions as shown in Figure~\ref{fig:contactresults}.
Additionally, the OpenPose confidence $c$ for each joint position is included as input.
Hence, the input to the network is a vector of $(x,y,c)$ values of dimension $3*13*9=351$.
The model outputs four contact labels (left/right toe, left/right heel) for a window of 5 frames centered around the target.  At test time, we use majority voting at overlapping predictions to smooth labels across time.

We use a five-layer multilayer perceptron (MLP) (sizes 1024, 512, 128, 32, 20) with ReLU non-linearities~\cite{pytorch}.
We train the network entirely on our synthetic dataset split 80/10/10 for train/validation/test based on motions per character, \ie, no motion will be in both train and test on the same character, but a training motion may appear in the test set retargeted to a different character. Although 3D motions may be similar in train and test, the resulting 2D motions (the network input) will be very different after projecting to differing camera viewpoints. The network is trained using a standard binary cross-entropy loss.

\subsection{Kinematic Initialization} 
\label{sec:kinematic}
Along with contact labels, our physics-based optimization requires as input a ground plane and initial trajectories for the COM, feet, and inertia tensor. In order to obtain these, we compute an initial 3D full-body motion from video.  Since this stage uses standard elements, e.g., \cite{Gleicher:1998:RMN}, we summarize the algorithm here, and provide full details in Appendix~\ref{appendix:kinematic}.

First, Monocular Total Capture \cite{mtc} (MTC) is applied to the input video to obtain an initial noisy 3D pose estimate for each frame. Although MTC accounts for motion through a texture-based refinement step, the output still contains a number of artifacts (Figure~\ref{fig:teaser}) that make it unsuitable for direct use in our physics optimization. 
Instead, we initialize a skeleton $\bS_{\textit{src}}$ containing 28 body joints from the MTC input poses, and then use a kinematic optimization to solve for an optimal root translation and joint angles over time, along with parameters of the ground plane. The objective for this optimization contains terms to smooth the motion, ensure feet are stationary and on the ground when in contact, and to stay close to both the 2D OpenPose and 3D MTC pose inputs. 

We first optimize so that the feet are stationary, but not at a consistent height. Next, we use a robust regression to find the ground plane which best fits the foot joint contact positions. Finally, we continue the optimization to ensure all feet are on this ground plane when in contact.

The full-body output motion of the kinematic optimization is used to extract inputs for the physics optimization. Using a predefined body mass (73 kg for all experiments) and distribution~\cite{DELEVA19961223}, we compute the COM and inertia tensor trajectories. We use the orientation about the root joint as the COM orientation, and the feet joint positions are used directly.

%% file: content/figures/pipeline.tex
\begin{figure*}[t]
\begin{center}
\includegraphics[width=0.95\linewidth]{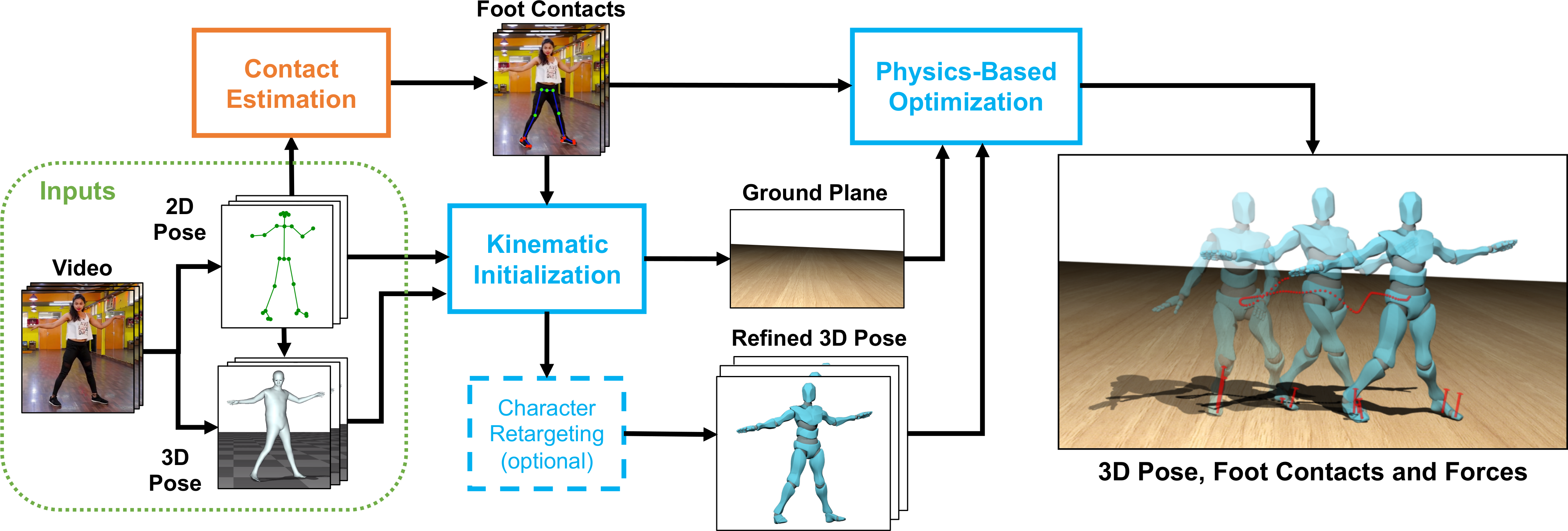}
\end{center}
  \caption{Method overview. Given an input video, our method starts with initial estimates from existing 2D and 3D pose methods \cite{cao2018openpose,mtc}. The lower-body 2D joints are used to infer foot contacts (orange box). Our optimization framework contains two parts (blue boxes). Inferred contacts and initial poses are used in a kinematic optimization that refines the 3D full-body motion and fits the ground. These are given to a reduced-dimensional physics-based trajectory optimization that applies dynamics.}
\label{fig:flowchart}
\end{figure*}

%% file: content/figures/physics_formulation.tex
\begin{figure}[t]
\begin{center}
\scriptsize{
\begin{align*}
    \min \mindent &  \sum\nolimits_{t=0}^{T} \Big( E_\textit{{data}}(t) + E_\textit{{vel}}(t) + E_\textit{{acc}}(t) \Big) + E_\textit{{dur}}\\
    \text{s.t.} \mindent &  m \ddot{\br}(t) = \sum\nolimits_{i=1}^{4} \beff_i(t) + m\bg \tag{dynamics}  \\
    & \bI_w(t) \dot{\bomega}(t) + \bomega(t) \times \bI_w(t) \bomega(t) = \sum\nolimits_{i=1}^{4} \beff_i(t) \times (\br(t) - \bp_i(t)) \\
    &  \dot{\br}(0) = \dot{\brb}(0), \dot{\br}(T) = \dot{\brb}(T) \tag{velocity boundaries} \\
    & || \bp_1(t) - \bp_2(t) || = || \bp_3(t) - \bp_4(t) || = \ell_\textit{{foot}} \tag{foot kinematics} \\
    & \text{ for every foot joint } i: \\
    & \mindent  || \bp_i(t) - \bp_{\textit{hip}, i}(t)|| \leq \ell_\textit{{leg}} \tag{leg kinematics} \\
    & \mindent \sum\nolimits_{j=1}^{n_{i}} T_{i, j} = T \tag{contact durations} \\
    & \text{ for foot joint } i \text{ in contact at time } t : \\
    & \mindent \dot{\bp}_i(t) = 0 \tag{no slip} \\
    & \mindent p_i^z (t) = h_\textit{{floor}} (\bp_i^{xy}) \tag{on floor} \\
    & \mindent 0 \leq \beff_i(t)^T \hat{\bn} \leq f_\textit{{max}} \tag{pushing/max force} \\
    & \mindent | \beff_i(t)^T \hat{\bt}_{1,2} | < \mu \beff_i(t)^T \hat{\bn} \tag{friction pyramid} \\
    & \text{ for foot joint } i \text{ in flight at time } t : \\
    & \mindent p_i^z (t) \geq h_\textit{{floor}} (\bp_i^{xy}) \tag{above floor} \\
    & \mindent \beff_i(t) = 0 \tag{no force in air}
\end{align*}
}
\end{center}
   \caption{Physics-based trajectory optimization formulation. Please see text for details.
 }
\label{fig:physicsformulation}
\end{figure}

%% file: content/text/04_experiments.tex
\section{Results}
Here we present extensive qualitative and quantitative evaluations of our contact estimation and motion optimization.

\subsection{Contact Estimation}\label{sec:contactres}
We evaluate our learned contact estimation method and compare to baselines on the synthetic test set (78 videos) and 9 real videos with manually-labeled foot contacts. The real videos contain dynamic dancing motions and include 700 labeled frames in total. 
In Table~\ref{table:contactestimation}, we report classification accuracy for our method and numerous baselines.

\input{content/tables/contact.tex}

\input{content/figures/qual_contact.tex}

We compare to using a velocity heuristic on foot joints, as described in Section~\ref{section:contact}, for both the 2D OpenPose and 3D MTC estimations. We also compare to using different subsets of joint positions. 
Our MLP using all lower-body joints is substantially more accurate on both synthetic and real videos than all baselines. Using upper-body joints down to the knees yields surprisingly good results.  

In order to test the benefit of contact estimation, we compared our full optimization pipeline on the synthetic test set using network-predicted contacts versus contacts predicted using a velocity heuristic on the 3D joints from MTC input. Optimization using network-predicted contacts converged for 94.9\% of the test set videos, compared to 69.2\% for the velocity heuristic. This illustrates how contact prediction is crucial to the success of motion optimization.

Qualitative results of our contact estimation method are shown in Figure~\ref{fig:contactresults}.
Our method is compared to the 2D velocity baseline which has difficulty for planted feet when detections are noisy, and often labels contacts for joints that are stationary but off the ground (\eg\ heels).

\input{content/figures/qual_motion.tex}

\subsection{Qualitative Motion Evaluation}\label{section:qualeval}
Our method provides key qualitative improvements over prior kinematic approaches. We urge the reader to \textbf{view the supplementary video} in order to fully appreciate the generated motions. For qualitative evaluation, we demonstrate animation from video by retargeting captured motion to a computer-animated character. Given a target skeleton $\bS_{\mathit{tgt}}$ for a character, we insert an IK retargeting step following the kinematic optimization as shown in Figure \ref{fig:flowchart} (see Appendix~\ref{appendix:retarget} for details), allowing us to perform the usual physics-based optimization on this new skeleton. We use the same IK procedure to compare to MTC results directly targeted to the character.

Figure \ref{fig:teaser} shows that our proposed method fixes artifacts such as foot floating (top row), foot penetrations (middle), and unnatural leaning (bottom). Figure \ref{fig:motionresults}(a) shows frames comparing the MTC input to our final result on a synthetic video for which we have a ground truth alternate view. For this example only, we use the true ground plane as input to our method for a fair comparison (see Section \ref{section:quanteval}). From the input view, our method fixes feet floating and penetration. From the first frame of the alternate view, we see that the MTC pose is in fact extremely unstable, leaning backward while balancing on its heels; our method has placed the contacting feet in a stable position to support the pose, better matching the true motion. 

Figure \ref{fig:motionresults}(b) shows additional qualitative results on a real video. We faithfully reconstruct dynamic motion with complex contact patterns in a physically accurate way. The bottom row shows the outputs of the physics-based optimization stage of our method at multiple frames: the COM trajectory and contact forces at the heel and toe of each foot.

\subsection{Quantitative Motion Evaluation}\label{section:quanteval}

Quantitative evaluation of high-quality motion estimation presents a significant challenge. 
Recent pose estimation work evaluates average positional errors of joints in the local body frame up to various global alignment methods \cite{Pavllo:2019:posemetric}.
However, those pose errors can be misleading: 
a motion can be pose-wise close to ground truth on average, but produce extremely implausible dynamics, including vibrating positions and extreme body lean. These errors can be perceptually objectionable when remapping the motion onto an animated character, and prevent the use of inferred dynamics for downstream vision tasks.

\input{content/figures/grf_examples.tex}

Therefore, we propose to use a set of metrics inspired by the biomechanics literature \cite{BrubakerICCV2009,HerrPopovic,Jiang:2019:SBR}, namely, to evaluate \textit{plausibility} of physical quantities based on known properties of human motion. 

We use two baselines: MTC, which is the state-of-the-art for pose estimation, and our kinematic-only initialization (Section \ref{sec:kinematic}), which transforms the MTC input to align with the estimated contacts from Section \ref{section:contact}. We run each method on the synthetic test set of 78 videos. For these quantitative evaluations only, we use the ground truth floor plane as input to our method to ensure a fair comparison. Note that our method does not \emph{need} the ground truth floor, but using it ensures a proper evaluation of our primary contributions rather than that of the floor fitting procedure, which is highly dependent on the quality of MTC input (see Appendix~\ref{appendix:results} for quantitative results using the estimated floor).
\paragraph{Dynamics Metrics.}
To evaluate dynamic plausibility, we estimate net ground reaction forces (GRF), defined as $\beff_{\mathit{GRF}}(t) = \sum_i \beff_i(t)$. For our full pipeline, we use the physics-based optimized GRFs which we compare to implied forces from the kinematic-only initialization and MTC input. In order to infer the GRFs implied by the kinematic optimization and MTC, we estimate the COM trajectory of the motion using the same mass and distribution as for our physics-based optimization (73 kg). We then approximate the acceleration at each time step and solve for the implied GRFs for all time steps (both in contact and flight).

We assess plausibility using GRFs measured in force plate studies, e.g., \cite{BrubakerICCV2009,HerrPopovic,RMB}. For walking, GRFs typically reach 80\% of body weight; for a dance jump, GRFs can reach up to about 400\% of body weight \cite{KuligGRF}. Since we do not know body weights of our subjects, we use a conservative range of 50kg--80kg for evaluation. Figure \ref{fig:grfexamples} shows the optimized GRFs produced by our method for a walking and swing dancing motion. The peak GRFs produced by our method match the data: for the walking motion, 115--184\% of body weight, and 127--204\% for dancing. In contrast, the kinematic-only GRFs are 319--510\% (walking) and 765--1223\% (dancing); these are implausibly high, a consequence of noisy and unrealistic joint accelerations.

\input{content/tables/perception.tex}

We also measure GRF plausibility across the whole test set (Table \ref{table:perceptioneval}(left)). GRF values are measured as a percentage of the GRF exerted by an idle 73 kg person. On average, our estimate is within 1\% of the idle force, while the kinematic motion implies GRFs as if the person were 24.4\% heavier. Similarly, the peak force of the kinematic motion is equivalent to the subject carrying an extra 830 kg of weight, compared to only 174 kg after physics optimization. The Max GRF for MTC is even less plausible, as the COM motion is jittery before smoothing during kinematic and dynamics optimization. \emph{Ballistic GRF} measures the median GRF on the COM when no feet joints should be in contact according to ground truth labels. The GRF should be exactly 0\%, meaning there are no contact forces and only gravity acts on the COM; the kinematic method obtains results of 255\%, as if the subject were wearing a powerful jet pack.
\paragraph{Kinematics Metrics.}
We consider three kinematic measures of plausibility (Table~\ref{table:perceptioneval}(right)). These metrics evaluate accuracy of foot contact measurements. Specifically, given ground truth labels of foot contact we compute instances of foot \emph{Floating}, \emph{Penetration}, and \emph{Skate} for heel and toe joints. \emph{Floating} is the fraction of foot joints more than 3 cm off the ground when they should be in contact. \emph{Penetration} is the fraction penetrating the ground more than 3 cm at any time. \emph{Skate} is the fraction moving more than 2 cm when in contact.

After our kinematics initialization, the scores on these metrics are best (lower is better for all metrics) and degrade slightly after adding physics. This is due to the IK step which produces full-body motion following the physics-based optimization. Both the kinematic and physics optimization results substantially outperform MTC, which is rarely at a consistent foot height.
\paragraph{Positional Metrics.}
For completeness, we evaluate the 3D pose output of our method on variations of standard positional metrics. Results are shown in Table~\ref{table:poseestimation}. In addition to our synthetic test set, we evaluate on all walking sequences from the training split of HumanEva-I \cite{Sigal:2010:HumanEva} using the known ground plane as input. We measure the mean \textbf{global} per-joint position error (mm) for ankle and toe joints (\textit{Feet} in Table~\ref{table:poseestimation}) and over all joints (\textit{Body}). We also report the error after aligning the root joint of only the first frame of each sequence to the ground truth skeleton (\textit{Body-Align 1}), essentially removing any spurious constant offset from the predicted trajectory. Note that this differs from the common practice of aligning the roots at every frame, since this would negate the effect of our trajectory optimization and thus does not provide an informative performance measure. The errors between all methods are comparable, showing at most a difference of 5 cm which is very small considering global joint position. Though the goal of our method is to improve physical plausibility, it does not negatively affect the pose on these standard measures.

\input{content/tables/pose.tex}

%% file: content/tables/contact.tex
\setlength{\tabcolsep}{4pt}
\begin{table}[t]
\caption{Classification accuracy of estimating foot contacts from video. Left: comparison to various baselines, Right: ablations using subsets of joints as input features.}
\begin{center}
\scalebox{0.75}{
\begin{tabular}{ r c c | r c c }
\toprule
 \textbf{Baseline} & \textbf{Synthetic} & \textbf{Real} & \multicolumn{1}{c}{\textbf{MLP}} & \textbf{Synthetic} & \textbf{Real} \\
 \multicolumn{1}{r}{\textbf{Method}} & \textbf{Accuracy} &  \textbf{Accuracy} & \multicolumn{1}{c}{\textbf{Input Joints}} & \textbf{Accuracy} & \textbf{Accuracy} \\
\midrule
Random & 0.507 & 0.480 & Upper down to hips & 0.919 & 0.692 \\
Always Contact & 0.677 & 0.647 & Upper down to knees & 0.935 & 0.865\\ 
2D Velocity & 0.853 &  0.867 & Lower up to ankles & 0.933 & 0.923 \\
3D Velocity & 0.818 & 0.875 & \textbf{Lower up to hips} & \textbf{0.941} & \textbf{0.935} \\ 
\bottomrule
\end{tabular}}
\end{center}
\label{table:contactestimation}
\end{table}

%% file: content/figures/qual_contact.tex
\begin{figure}[t]
\begin{center}
   \includegraphics[width=0.7\linewidth]{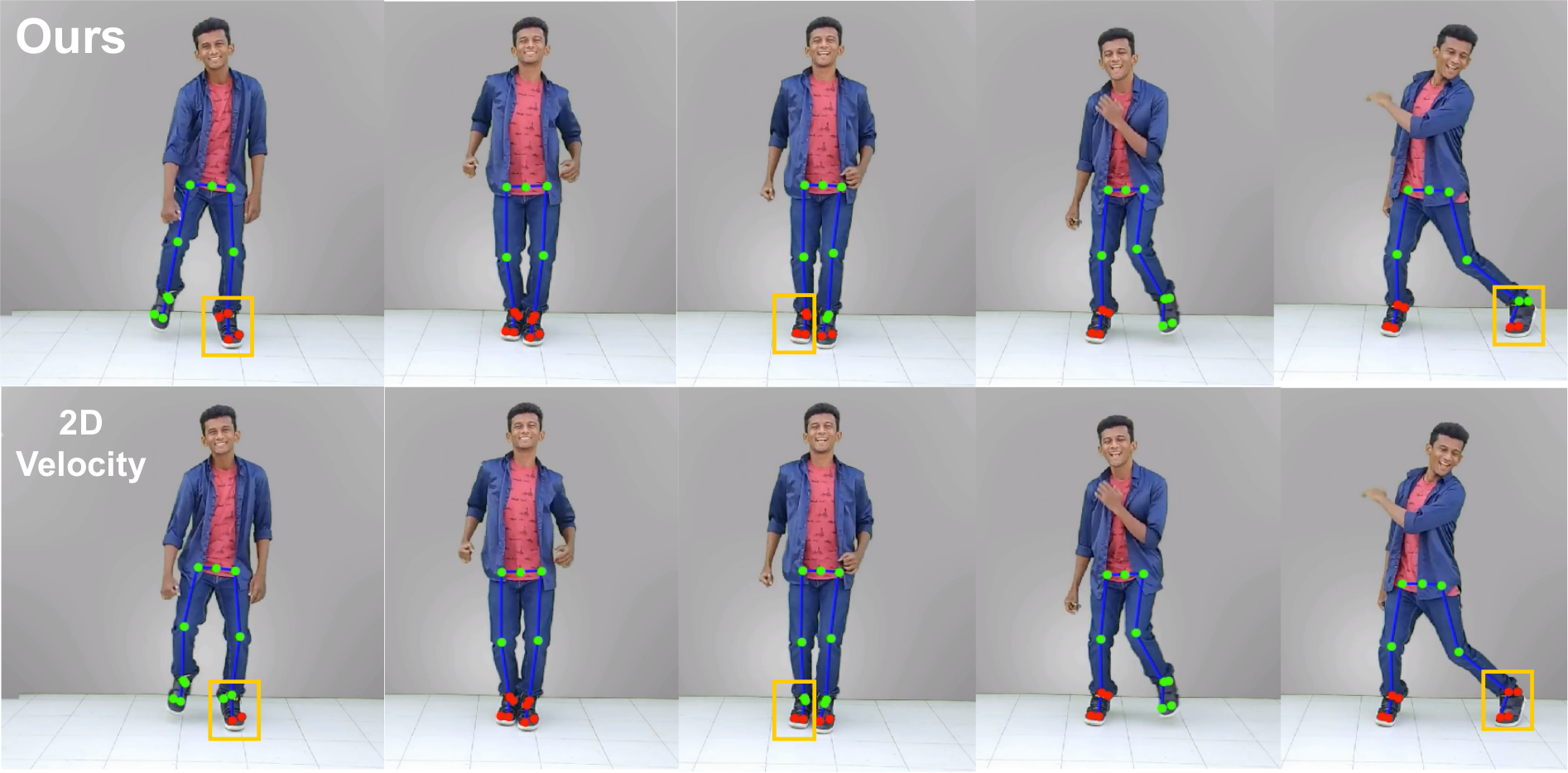}
\end{center}
   \caption{Foot contact estimation on a video using our learned model compared to a 2D velocity heuristic. All visualized joints are used as input to the network which outputs four contact labels (left toes, left heel, right toes, right heel). Red joints are labeled as contacting. Key differences are shown with orange boxes.
 }
\label{fig:contactresults}
\end{figure}

%% file: content/figures/qual_motion.tex
\begin{figure*}
\begin{center}
\includegraphics[width=0.99\linewidth]{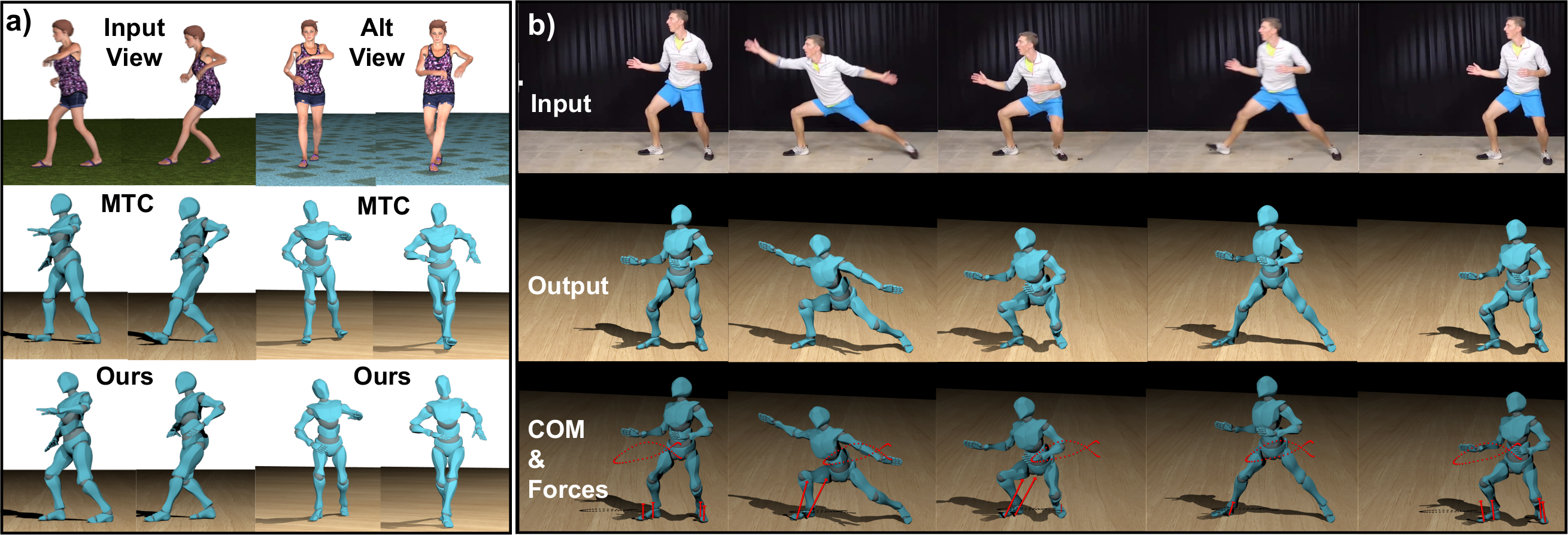}
\end{center}
  \caption{Qualitative results on synthetic and real data. a) results on a synthetic test video with a ground truth alternate view. Two nearby frames are shown for the input video and the alternate view. We fix penetration, floating and leaning prevalent in our method's input from MTC. b) dynamic exercise video (top) and the output full-body motion (middle) and optimized COM trajectory and contact forces (bottom).}
\label{fig:motionresults}
\end{figure*}

%% file: content/figures/grf_examples.tex
\begin{wrapfigure}{r}{0.5\textwidth}
\centering
   \includegraphics[width=0.5\textwidth]{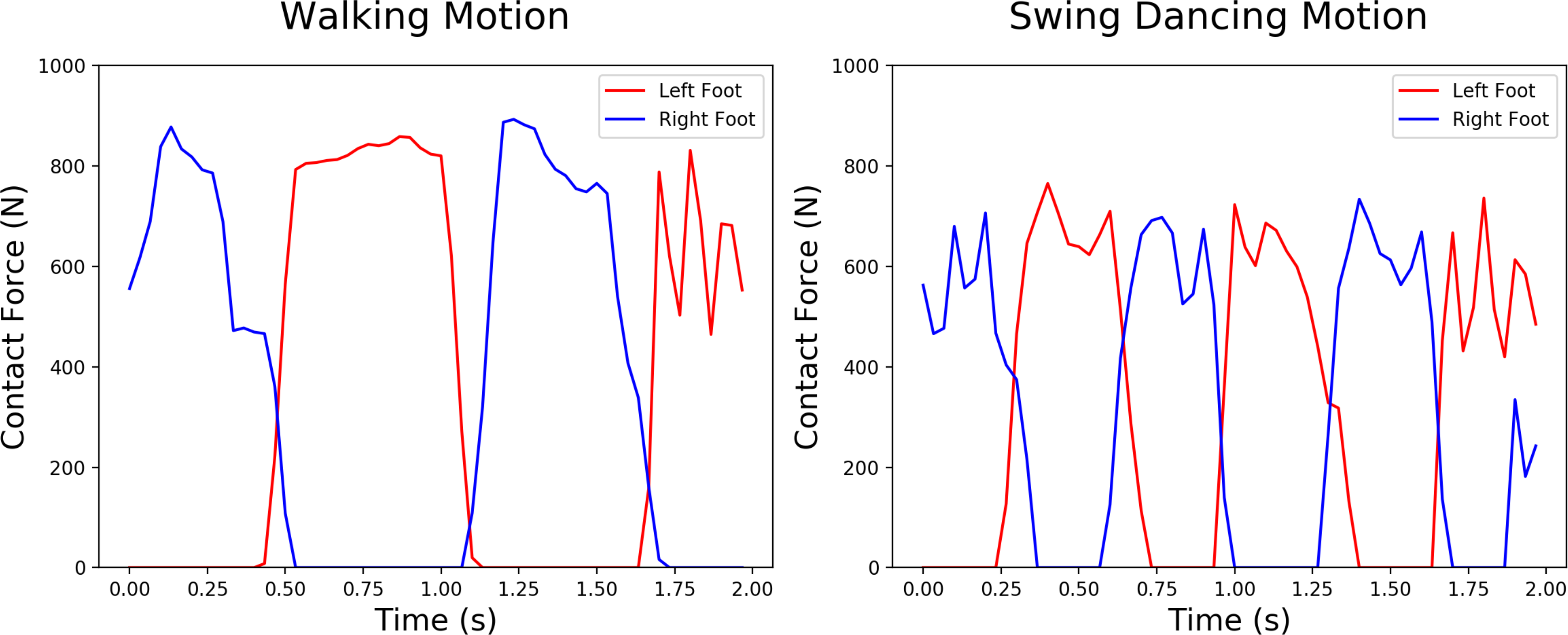}
   \caption{Contact forces from our physics-based optimization for a walking and dancing motion. The net contact forces around 1000 N are 140\% of the assumed body weight (73 kg), a reasonable estimate compared to prior force plate data \cite{BrubakerICCV2009}.
 }
\label{fig:grfexamples}
\vspace{-15pt}
\end{wrapfigure}

%% file: content/tables/perception.tex
\setlength{\tabcolsep}{4pt}

\begin{table*}[t]
\caption{Physical plausibility evaluation on synthetic test set. \emph{Mean/Max GRF} are contact forces as a proportion of body weight; see text for discussion of plausible values. \emph{Ballistic GRF} are unexplained forces during flight; smaller values are better. Foot position metrics measure the percentage of frames containing typical foot contact errors per joint; smaller values are better.}
\begin{center}
\scalebox{0.8}{
\begin{tabular}{ r | r r r | r r r }
\toprule
  & \multicolumn{3}{c}{Dynamics (Contact forces)} & \multicolumn{3}{c}{Kinematics (Foot positions)}  \\
\midrule
 \textbf{Method} & \textbf{Mean GRF} & \textbf{Max GRF} & \textbf{Ballistic GRF} & \textbf{Floating} & \textbf{Penetration} & \textbf{Skate} \\
\midrule
MTC \cite{mtc} & 143.0\% & 9055.3\% & 115.6\% & 58.7\% &	21.1\% &	16.8\%\\
Kinematics (ours) &  124.4\% &	1237.5\% & 255.2\% & \textbf{2.3\%} &	2.8\% &	\textbf{1.6\%} \\
Physics (ours) & \textbf{99.0\%} &\textbf{338.6\%} & \textbf{0.0\%} & 8.2\% & \textbf{0.3\%} &	3.6\% \\
\bottomrule
\end{tabular}}
\end{center}
\label{table:perceptioneval}
\end{table*}

%% file: content/tables/pose.tex
\setlength{\tabcolsep}{4pt}

\begin{table}[t]
\caption{Pose evaluation on synthetic and HumanEva-I walking datasets. We measure mean global per-joint 3D position error (no alignment) for feet and full-body joints. For full-body joints, we also report errors after root alignment on only the first frame of each sequence. We remain competitive while providing key physical improvements.}
\begin{center}
\scalebox{0.8}{
\begin{tabular}{ r | c c c | c c c }
\toprule
& \multicolumn{3}{c}{Synthetic Data} & \multicolumn{3}{c}{HumanEva-I Walking} \\
\midrule
 \textbf{Method} & \textbf{Feet}  & \textbf{Body} &  \textbf{Body-Align 1} & \textbf{Feet} & \textbf{Body} &  \textbf{Body-Align 1} \\
\midrule
MTC \cite{mtc} &  581.095 & \textbf{560.090} & \textbf{277.215} &  511.59 & 532.286 & \textbf{402.749} \\ 
Kinematics (ours) 	& 573.097 & 562.356	& 281.044	& \textbf{496.671}	& 525.332	& 407.869 \\
Physics (ours) &  \textbf{571.804} & 573.803 & 323.232 &	508.744	& \textbf{499.771} & 421.931 \\
\bottomrule
\end{tabular}}
\end{center}
\label{table:poseestimation}
\end{table}

%% file: content/text/05_discussion.tex
\section{Discussion}
\paragraph{Contributions.}
The method described in this paper estimates physically-valid motions from initial kinematic pose estimates.  As we show, this produces motions that are visually and physically much more plausible than the state-of-the-art methods.  We show results on retargeting to characters, but it could also be used for further vision tasks that would benefit from dynamical properties of motion.

Estimating accurate human motion entails numerous challenges, and we have focused on one crucial sub-problem. There are several other important unknowns in this space, such as motion for partially-occluded individuals, and ground plane position. Each of these problems and the limitations discussed below are an enormous challenge in their own right and are therefore reserved for future work. However, we believe that the ideas in this work could contribute to solving these problems and open multiple avenues for future exploration.
\paragraph{Limitations.}
We make a number of assumptions to keep the problem manageable, all of which can be relaxed in future work: we assume that feet are unoccluded, there is a single ground plane, the subject is not interacting with other objects, and we do not handle contact from other body parts like knees or hands. These assumptions are permissible for the character animation from video mocap application, but should be considered in a general motion estimation approach.
Our optimization is expensive. For a 2 second (60 frame) video clip, the physical optimization usually takes from 30 minutes to 1 hour. This runtime is due primarily to the adapted implementation from prior work~\cite{winkler18} being ill-suited for the increased size and complexity of human motion optimization. We expect a specialized solver and optimized implementation to speed up execution.

%% file: appendix/macros.tex
\newcounter{appendixsection}
\setcounter{appendixsection}{0}
\newcounter{appendixfigure}
\setcounter{appendixfigure}{0}
\newcounter{appendixtable}
\setcounter{appendixtable}{0}
\newcounter{appendixequation}
\setcounter{appendixequation}{0}

\renewcommand\thesection{\Alph{appendixsection}}
\renewcommand\thesubsection{\Alph{appendixsection}.\arabic{subsection}}
\renewcommand\thefigure{\Alph{appendixsection}\arabic{appendixfigure}}    
\renewcommand\thetable{\Alph{appendixsection}\arabic{appendixtable}}    
\renewcommand\theequation{\Alph{appendixsection}\arabic{appendixequation}}

\newcommand{\bqb}{\overline{\bq}}

%% file: appendix/text/02_contact.tex
\stepcounter{appendixsection}
\setcounter{appendixfigure}{0}
\setcounter{appendixtable}{0}
\setcounter{appendixequation}{0}
\section{Contact Estimation Details}
\label{appendix:contact}

Here we detail the contact estimation model and the dataset from Section~\ref{section:contact}.

\subsection{Synthetic Dataset}
Our synthetic dataset was rendered using Blender\footnote{https://www.blender.org/} and includes 13 characters performing 65 different motion capture sequences retargeted to each character taken from \url{www.mixamo.com}. Each motion is recorded from 2 camera viewpoints resulting in 1690 videos and 101k frames of data. The motions include: samba, swing, and salsa dancing, boxing, football, and baseball actions, walking, and idle poses. Videos are rendered at 1280x720 with motion blur, and are 2 seconds long at 30 fps. Example frames from the dataset are shown in Figure~\ref{fig:data}. In addition to RGB frames, at each timestep the dataset includes 2D OpenPose~\cite{cao2018openpose} detections, the 3D pose in the form of the character's skeleton (skeletons are different for each character, and pose is provided in a .bvh motion capture file), foot contact labels for the heel and toe base of each foot as described in the main paper, and camera parameters.

For each video, many parameters are randomized. The camera is placed at a uniform random distance in $[4.5, 7.5]$ and Gaussian random height with $\mu = 0.9$, and $\sigma = 0.3$ but clamped to be in $[0.3, 1.75]$, all in meters. The camera is placed at a random angle within 90 degrees offset from the front of the character but always looks at roughly character hip height. The camera does not move during the video. Floor texture is randomly chosen from solid colors and 26 other textures with various wood, grass, tile, metal, and carpet. Four lights in the scene are randomly toggled on and off, given random energies, and randomly offset from default positions, resulting in many shadow variations across videos. 

\input{appendix/figures/data_frames.tex}

\subsection{Model Details}
We implement our contact estimation MLP (sizes 1024, 512, 128, 32, 20) in PyTorch \cite{pytorch}. All but the last layer are followed by batch normalization, and we use a single dropout layer before the size-128 layer (dropout $p=0.3$). To train, we optimize the binary cross-entropy loss using Adam \cite{adam} with a learning rate of $10^{-4}$. We apply an L2 weight decay with a weight of $10^{-4}$ and use early stopping based on the validation set loss. We scale all 2D joint inputs to be largely between $[-1, 1]$. During training, we also add Gaussian noise to the normalized joints with $\sigma=0.005$.

Because our network classifies contacts jointly over 5 frames for every target frame (the frame at the center of the window), there are many overlapping predictions at test time. When inferring contacts for an entire video at test time, we first use every frame as a target and then collect votes from overlapping predictions. A joint is marked in contact at a frame if a majority of the votes for that frame classify it as in contact.

%% file: appendix/figures/data_frames.tex
\stepcounter{appendixfigure}
\begin{figure}[t]
\begin{center}
\includegraphics[width=1.0\linewidth]{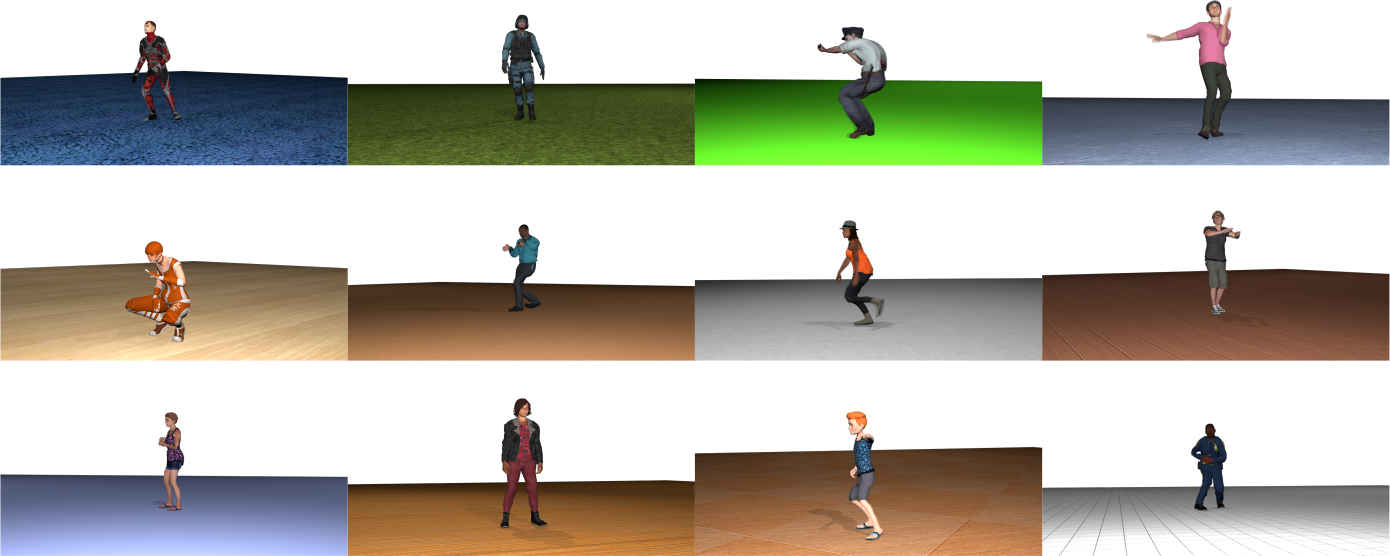}
\end{center}
   \caption{Example RGB frames from our synthetic dataset for contact estimation learning. The data also contains 2D OpenPose~\cite{cao2018openpose} detections, 3D pose in the form of the character's skeleton, and automatically labeled contacts for the toe base and heels.
   }
\label{fig:data}
\end{figure}

%% file: appendix/text/03_kinematic.tex
\input{appendix/figures/skel.tex}
\stepcounter{appendixsection}
\setcounter{appendixfigure}{0}
\setcounter{appendixtable}{0}
\setcounter{appendixequation}{0}
\section{Kinematic Optimization Details}
\label{appendix:kinematic}

Here detail the kinematic optimization procedure used to initialize our physics-based optimization. See Section~\ref{sec:kinematic} for an overview. 

\subsection{Inputs and Initialization}
\label{appendix:kinematic:sec:inputs}

Our kinematic optimization takes in $J$ body joints over $T$ timesteps that make up the motion. Specifically, for $j\in J$ and $t\in T$ we have the 2D pose detection from OpenPose $\bx_{j,t} \in \reals^2$ with confidence $\sigma_{j,t}$, contacts estimated in the previous stage of our pipeline $c_{j,t} \in \{0, 1\}$ where $c_{j,t}=1$ if joint $j$ is in contact with the ground at time $t$, and finally the full-body 3D pose from MTC. 

We use the MTC input to initialize a custom skeleton, called $\bS_{\textit{src}}$ in the main paper, which contains $J=28$ joints. In particular, we use the MTC COCO regressor to obtain 19 body joint positions (excluding feet) over the sequence, and vertices on the Adam mesh for the 6 foot joints as in the original MTC paper. We choose these 25 joints in order to use a re-projection error in our optimization objective, as described below. To better map to the character rigs used during animation retargeting, we additionally use 3 spine joint positions directly from the MTC body model giving the final total of 28 joints. Note, we do not use any hand or face information from MTC. Our skeleton has fixed bone lengths which are determined based on these input 3D joint locations throughout the motion sequence: the median distance between two joints over the entire motion sequence defines the bone length. Our skeleton (before fitting bone lengths to the input data) is visualized in Figure~\ref{fig:skel}.

We normalize the input positions to get the root-relative positions of each joint $\bqb_{j,t} \in \mathbb{R}^3, \; j = 1,\dots,J, \; t = 1,\dots,T$ which we will target during optimization, and let the global translation be our initial root translation $\bpb_{\textit{root},t}$. All these positions are preprocessed to remove obviously incorrect frames based on OpenPose detection confidence: for the 25 joints with corresponding 2D OpenPose detections (all non-spine joints in our skeleton), if the confidence is below 0.3, then the frame is replaced with a linear interpolation between the closest adjacent frames with sufficient confidence. 

Because we optimize for the joint angles of our skeleton (see below), next we must find initial joint angles to match the MTC joint position inputs. We roughly initialize the joint angles of our skeleton by copying those from the MTC body model, and finally perform inverse kinematics (IK) targeting the preprocessed joint positions which results in a reconstruction of the MTC input on our skeleton. We use a Jacobian-based full body IK solver based on \cite{Choi:1999:OMR}. This is the skeleton which is optimized throughout our kinematic initialization.

\subsection{Optimization Variables}
We optimize over global 3D root translation $\bp_{\textit{root},t} \in \reals^3$ and skeleton joint Euler angles $\btheta_{j,t} \in \reals^3$ with $j = 1,\dots,J, \; t = 1,\dots,T$. We also find ground plane parameters $\hat{\mathbf{n}}, \bp_{\textit{floor}} \in \mathbb{R}^3$ which are the normal vector and some point on the plane. As described below, we do not jointly optimize all of these at once; we do it in stages and fit the floor separately.

\subsection{Problem Formulation}
We seek to minimize the following objective function:
\begin{equation*}
     \alpha_{\textit{proj}} \E{\textit{proj}} + \alpha_{\textit{vel}} \E{\textit{vel}} + \alpha_{\textit{ang}} \E{\textit{ang}} + \alpha_{\textit{acc}} \E{\textit{acc}} + \alpha_{\textit{data}} \E{\textit{data}} + \alpha_{\textit{cont}} \E{\textit{cont}} + \alpha_{\textit{floor}} \E{\textit{floor}}
\end{equation*}
where the $\alpha$ are constant weights. We use $\alpha_{\textit{proj}} = 0.5$, $\alpha_{\textit{vel}}=\alpha_{\textit{ang}}=0.1$, $\alpha_{\textit{acc}}=0.5$, $\alpha_{\textit{data}}=0.3$, and $\alpha_{\textit{cont}}=\alpha_{\textit{floor}}=10$.
We now detail each of these energy terms. 

Suppose $\bq_{j,t} \in \mathbb{R}^3, \; j = 1,\dots,J, \; t = 1,\dots,T$ are the current root-relative joint position estimates during optimization which can be calculated using forward kinematics on $\bS_{\textit{src}}$ with the current joint angles $\btheta_{j,t}$. Then our energy terms are defined as follows.

\begin{itemize}
    \item \emph{2D Re-projection Error}: minimizes deviation of joints from corresponding OpenPose detections, weighted by detection confidence
    \stepcounter{appendixequation}
    \begin{equation}
        \E{\textit{proj}} = \sum_{t = 1}^T \sum_{j = 1}^J \sigma_{j,t} || \Pi (\bq_{j,t} + \bp_{\textit{root},t}) - \bx_{j,t} ||^2
    \end{equation}
    where $\Pi$ is the perspective projection parameterized by focal length $f$ (assumed to be 2000) and $[c_x, c_y]$.
    \item \emph{Velocity Smoothing}: minimizes change in joint positions and angles over time
    \stepcounter{appendixequation}
    \begin{align}
        \E{\textit{vel}} &= \sum_{t = 1}^{T-1} \sum_{j = 1}^J || \bq_{j, t+1} - \bq_{j,t} ||^2 \\
        \stepcounter{appendixequation}
        \E{\textit{ang}} &= \sum_{t = 1}^{T-1} \sum_{j = 1}^J || \btheta_{j,t+1} - \btheta_{j,t} ||^2.
    \end{align}
    \item \emph{Linear Acceleration Smoothing}: minimizes change in joint linear velocity over time
    \stepcounter{appendixequation}
    \begin{equation}
        \E{\textit{acc}} = \sum_{t = 1}^{T-2} \sum_{j = 1}^J || (\bq_{j, t+2} - \bq_{j, t+1}) - (\bq_{j,t+1} - \bq_{j,t}) ||^2.
    \end{equation}
    \item \emph{3D Data Error}: minimizes deviation from 3D MTC joint initialization
    \stepcounter{appendixequation}
    \begin{equation}
        \E{\textit{data}} = \sum_{t = 1}^T \sum_{j = 1}^J || \bq_{j,t} - \bqb_{j,t} ||^2.
    \end{equation}
    \item \emph{Contact Velocity Error}: encourages feet joints (toes and heels) to be stationary when labeled as in contact
    \stepcounter{appendixequation}
    \begin{equation}
        \E{\textit{cont}} =  \sum_{t = 1}^{T-1} \sum_{j \in J_F} || c_{j,t} \left((\bq_{j,t+1} + \bp_{\textit{root},t+1}) - (\bq_{j,t} + \bp_{\textit{root},t}) \right) ||^2.
    \end{equation}
    where $J_F$ is the set of foot joints.
    \item \emph{Contact Position Error}: encourages toe and heel joints to be on the ground plane when labeled as in contact
    \stepcounter{appendixequation}
    \begin{equation}
        \E{\textit{floor}} = \sum_{t = 1}^{T} \sum_{j \in J_F} || c_{j,t} \left(\hat{\mathbf{n}} \cdot (\bq_{j,t} + \bp_{\textit{root},t} - \bp_\textit{floor}) \right) ||^2.
    \end{equation}
\end{itemize}

\subsection{Optimization Algorithm}
We perform this optimization in three main stages. First, we enforce all objectives \textit{except} the contact position error and solve only for skeleton root position and joint angles (no floor parameters). Next, we use a \href{https://scikit-learn.org/stable/modules/linear_model.html#huber-regression}{robust Huber regression} to find the floor plane that best matches the foot joint contact positions and reject outliers, i.e., joints labeled as in contact when they are far from the ground. Outlier contacts are re-labeled as non-contacts for all subsequent processing. Finally, we repeat the full-body optimization, now enabling the contact position objective to ensure feet are on the ground plane. We optimize using the Trust Region Reflective algorithm with analytical derivatives.

\subsection{Extracting Inputs for Physics-Based Optimization}
From the full-body output of this kinematic optimization, we need to extract inputs for the physics-based optimization (Section~\ref{section:physicsoptim}). To get the COM targets $\brb(t) \in \reals^3$, we treat each body part as a point with a pre-defined mass \cite{DELEVA19961223}. This also allows the calculation of the body-frame inertia tensor at each time step $\bI_b(t) \in \reals^{3\times3}$ which is used to enforce dynamics constraints. Unless otherwise noted, we assume a body mass of 73 kg for the character. We use the orientation about the root joint as the COM orientation $\bthetab(t) \in \reals^3$ and the feet joint positions $\bpb_{1:4}(t) \in \reals^3$ are directly taken from the full-body motion.

%% file: appendix/figures/skel.tex
\stepcounter{appendixfigure}
\begin{wrapfigure}{r}{0.3\textwidth}
\centering
\vspace{-20pt}
\includegraphics[width=0.3\textwidth]{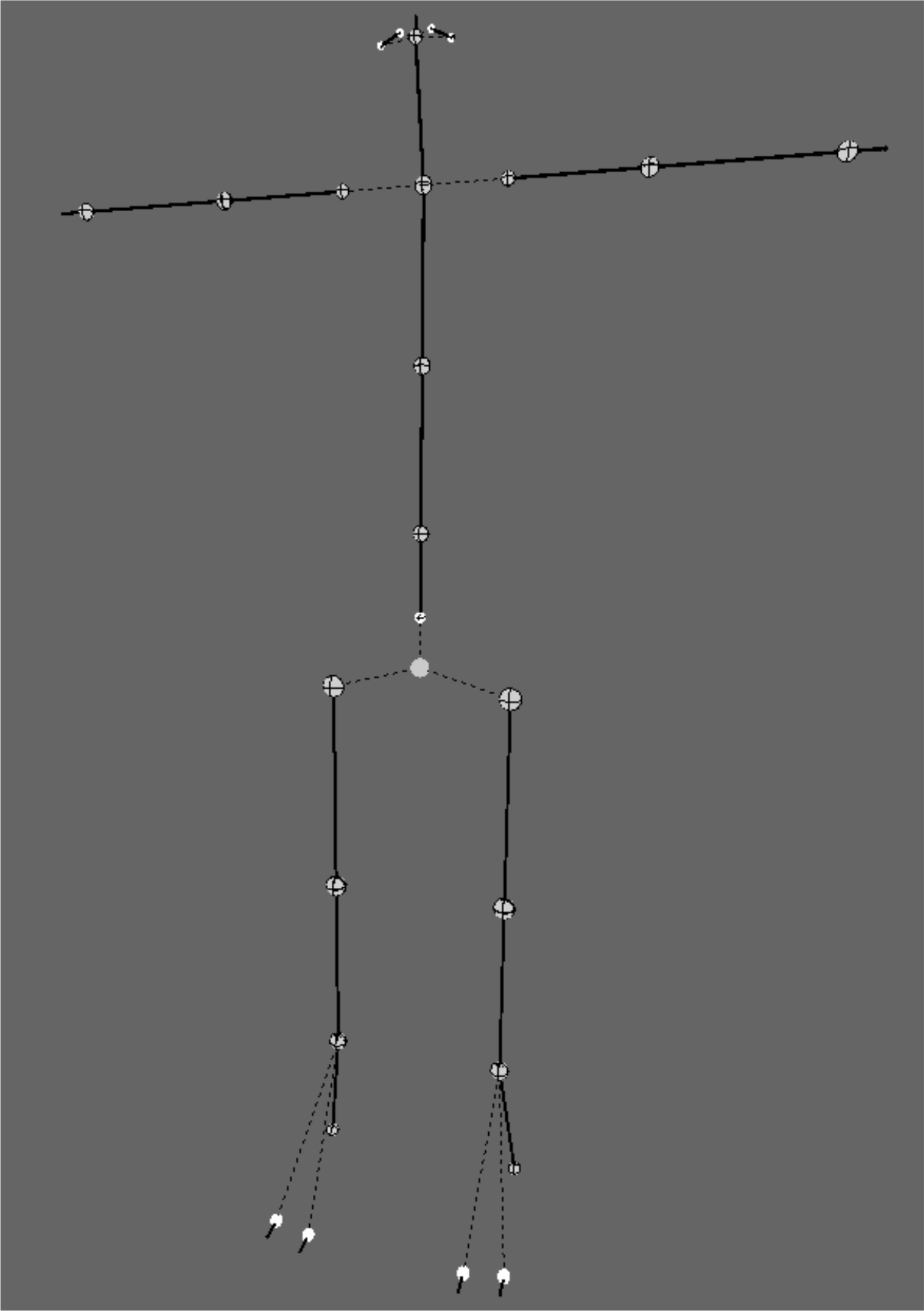}
\caption{Skeleton. Bone lengths and pose are initialized from MTC input for each motion sequence before kinematic optimization.}
\vspace{-40pt}
\label{fig:skel}
\end{wrapfigure}

%% file: appendix/text/05_physics_optim.tex
\stepcounter{appendixsection}
\setcounter{appendixfigure}{0}
\setcounter{appendixtable}{0}
\setcounter{appendixequation}{0}
\section{Physics-Based Optimization Details}
\label{appendix:physoptim}

Here we detail the physics-based trajectory optimization from Section~\ref{section:physicsoptim}.

\subsection{Polynomial Parameterization}
COM position and orientation, foot positions during flight, and contact forces during stance are parameterized by a sequence of cubic polynomials as done in Winkler et al.~\cite{winkler18}. These polynomials use a Hermite parameterization: we do not optimize over the polynomial coefficients directly, rather the duration, starting and ending positions, and boundary velocities.

The COM position and orientation use one polynomial every 0.1 s. Feet positions and forces always use at least 6 polynomials per phase, which we found necessary to accurately produce extremely dynamic motions. We adaptively add polynomials depending on the length of the phase. If the phase is longer than 2 s, additional polynomials are added commensurately. Foot positions during stance are a single constant value and contact forces during flight are constant 0 value. This ensures that the no slip and no force during flight constraints are met. 

Please see Winkler et al. \cite{winkler18} for a more in-depth discussion of the polynomial parameterization along with the contact phase duration parameterization.

\subsection{Constraint Parameters}
Though the optimization variables are continuous polynomials, objective energies and constraints are enforced at discrete intervals. Leg and foot kinematic constraints are enforced at 0.08 s intervals, the above floor constraint at 0.1 s intervals, and dynamics constraints are enforced every 0.1 s. In practice, the velocity boundary constraints try to match the \emph{mean} initial velocity over the first(last) 5 frames to make it more robust to noisy motion.

Objective terms, including smoothing, are enforced at every step for which we have input data. For example, the synthetic dataset at 30 fps will provide an objective term at $(1/30)$ s intervals.

\subsection{Contact Timing Optimization}
As explained in Section~\ref{section:physicsoptim}, our physics optimization is done in stages such that contact phase durations are not optimized until the very last stage. We found that allowing these durations to be optimized along with dynamics does not always result in a better solution as it is an inherently harder and less stable optimization. Therefore, in the presented results we use the better of the two solutions: either the solution using fixed input contact timings (from our neural network) or the solution after subsequently allowing phase durations to change, if the motion is improved.

\subsection{Full-Body Output}
Following the physics-based optimization, we must compute a full-body motion from the physically-valid COM and foot joint positions using IK. For the upper body (including the root), we calculate the offset of each joint from the COM in the input motion to the physics optimization, and use this offset added to the new optimal COM as the joint targets during IK. This means the upper-body motion will be essentially identical to the result of the kinematic optimization (though the posture may improve due to the new COM position). For the lower body, we target the toe and heel joints directly to the physically optimized output and let the remainder of the joints (i.e., ankles, knees, and hips) result from IK, which can be drastically different from the input. We use the same IK algorithm as in Appendix~\ref{appendix:kinematic:sec:inputs}.

%% file: appendix/text/04_retargeting.tex
\stepcounter{appendixsection}
\setcounter{appendixfigure}{0}
\setcounter{appendixtable}{0}
\setcounter{appendixequation}{0}
\section{Retargeting to a New Character}
\label{appendix:retarget}

In many cases, we wish to retarget the estimated motion to a new animated character mesh. We do this in the main paper in Section~\ref{section:qualeval} for qualitative evaluation. One could apply physics-based motion retargeting methods to the output motion after an IK retargeting procedure, e.g., \cite{Popovic:1999:PBM}. However, we avoid this extra step by directly performing our physics-based optimization on the target character skeleton.

Given a target skeleton $\bS_{\mathit{tgt}}$, we insert an additional retargeting step following the kinematic optimization (see Figure~\ref{fig:flowchart}). Namely, we uniformly scale $\bS_{src}$ to the approximate size of our target skeleton, and then perform an IK optimization based on a predefined joint mapping to recover joint angles for $\bS_{tgt}$. Then, the subsequent physics-based optimization and full-body upgrade are performed with this skeleton replacing $\bS_{src}$. We use the same IK algorithm as in Appendix~\ref{appendix:kinematic:sec:inputs}.

Note for qualitative comparison to MTC, we perform a very similar procedure: we first fit our skeleton to the raw MTC input, similar to Appendix~\ref{appendix:kinematic:sec:inputs} but without the preprocessing, and then perform the IK retargeting as described in this section. This provides a stronger baseline than a naive approach like directly copying joint angles from MTC to $\bS_{\mathit{tgt}}$.

%% file: appendix/text/07_results.tex
\stepcounter{appendixsection}
\setcounter{appendixfigure}{0}
\setcounter{appendixtable}{0}
\setcounter{appendixequation}{0}
\section{Evaluation Details and Additional Results}
\label{appendix:results}
Here we include additional results and details for evaluations.

\input{appendix/tables/contact_f1}
\input{appendix/tables/contact_window}

\subsection{Contact Estimation}
Main results for contact estimation are presented in Section~\ref{sec:contactres}. Table~\ref{table:contactf1} supplements the main results with the precision, recall, and F1 score of each method. These give additional insight compared to accuracy since data labels are slightly imbalanced (more in-contact frames than no-contact).

Table~\ref{table:windowablation} shows an ablation study between different input and output window size combinations for our network. \emph{Input Window} is the number of frames of 2D lower-body joints given to the network, and \emph{Prediction Window} is the number of frames for which the network outputs foot contact classifications. We use an input window $w = 9$ and prediction window of 5 in our experiments since it achieves the best accuracy on real videos as shown in the table. In general, there is not a clear trend in \emph{Prediction Window} size, but as the \emph{Input Window} size increases, so does accuracy on the real dataset.

\input{appendix/figures/offtarget.tex}

Figure~\ref{fig:offtarget} shows the accuracy of contact estimations over the entire prediction window of 5 frames on the synthetic test set. Though the target frame in this case is frame index 2, predictions on the off-target frames degrade only slightly and are still very accurate since the input windows is 9 frames. This motivates the use of the majority voting scheme at inference time.

\subsection{Qualitative Motion Evaluation}
For extensive qualitative evaluation, see the supplementary video. For real data, we use videos from: publicly available datasets \cite{chan2019dance,Peng:2018:SRL}, YouTube videos that are licensed under Creative Commons or with permission from the content creators to be used in this publication, and licensed stock footage.

\vspace{-5pt}
\subsection{Quantitative Motion Evaluation}
Primary quantitative results for motion reconstruction are presented in Section~\ref{section:quanteval}. These quantitative evaluations make use of the ground truth floor plane as input. Note that our method does not \emph{need} the ground truth floor: our floor fitting procedure works well as demonstrated in all qualitative results on live action monocular videos. However, we performed quantitative evaluations on data that contains many cases of movement directly towards or away from the camera: a challenging case for MTC, which results in noisy feet joints as input to our method causing a poor floor fit. This makes optimization difficult and interferes with evaluating our primary contributions.

However, for completeness, here we include quantitative results using the floor fitting procedure (rather than taking the ground truth floor as input) on the synthetic test set. Table \ref{table:fittedperceptioneval} shows kinematic and dynamics evaluations using the fitted floor while Table \ref{table:fittedposeestimation} shows the pose evaluation. Trends are similar to those in Tables~\ref{table:perceptioneval} and \ref{table:poseestimation} using the ground truth floor.

\vspace{-5pt}
\subsection{Discussion on Global Pose Estimation}
For quantitative evaluations in Section~\ref{section:quanteval}, we do not compare to other methods in global human motion estimation but instead evaluate ablations of our own method: the kinematics-only version and initialization from MTC. \input{appendix/tables/fitted_perception}
\input{appendix/tables/fitted_pose}
The problem of predicting a temporally-consistent global motion (like MTC and this work does) is vastly underexplored so there are few comparable prior works. Many methods do traditional local 3D pose estimation or even predict the global root translation from the camera, but these rarely result in a coherent global motion.

For example, a recent work from Pavlakos et al.~\cite{SMPL-X:2019} called SMPLify-X estimates global camera extrinsics, local pose, and body shape, which gives global motion when applied to video. However, we found that MTC, which uses a temporal tracking procedure, gave better results which motivated its use in initializing our pipeline. Figure~\ref{fig:mtcsmplifycompare} shows a fixed side view of results from SMPLify-X and MTC on the same video clip. SMPLify-X is noisy and inconsistent especially in terms of global translation; MTC is much smoother and coherent.

\subsection{Pose Estimation Evaluation Details}
We quantitatively evaluate pose estimation in Section~\ref{section:quanteval}. We evaluate on our synthetic test set and HumanEva-I~\cite{Sigal:2010:HumanEva} walking sequences. Like many pose estimation benchmarks (\eg~Human3.6M~\cite{h36m_pami}), few motions in HumanEva are dynamic with interesting foot contact patterns. Therefore, we evaluate on a subset containing the walking sequences which meet this criteria.

For the MTC baseline, we measure accuracy based directly on the regressed joints given as input to our method. For our method, we use the estimated joints after the full physics-based motion pipeline on our custom skeleton that is initially fit from the MTC input as described in Appendix~\ref{appendix:kinematic:sec:inputs}.

\input{appendix/figures/mtc_compare.tex}

For the synthetic test set, we measure joint errors with respect to a subset of the known character rig that includes 16 joints: neck, shoulders, elbows, wrists, hips, knees, ankles, and toes (no spine joints). The ``Feet" column of Table~\ref{table:poseestimation} includes ankle and toe joints only. 

On the right side of Table~\ref{table:poseestimation}, we evaluate methods on the walking sequences from the training split of HumanEva-I \cite{Sigal:2010:HumanEva} (which includes subjects 1, 2, and 3). Following prior work~\cite{Pavllo:2019:posemetric}, we first split the walking sequences into contiguous chunks by removing corrupted motion capture frames. We then further split these chunks into sequences of roughly 120 frames (about 2 seconds) to use as input to our method. We extract the ground truth floor plane using the camera extrinsics from the dataset and use this as input to our method. Joint errors are measured with respect to an adapted 15-joint skeleton \cite{Pavllo:2019:posemetric} from the HumanEva ground truth motion capture which includes: root, head, neck, shoulders, elbows, wrists, hips, knees, and ankles. The ``Feet" column of  Table~\ref{table:poseestimation} includes ankle joints only.

\input{appendix/figures/multi_char.tex}

\subsection{Multi-Character Generalization}
Following the procedure laid out in Appendix~\ref{appendix:retarget}, our physics-based optimization can be applied to many character skeletons with varying body and mass distributions. Figure~\ref{fig:chars} shows an example of estimating motion from the same video for three different characters: Ybot, Ty, and Skeleton Zombie. Ybot has a body mass of 73 kg with a typical human mass distribution \cite{DELEVA19961223}. Ty is much lighter at 36.5 kg and his distribution is modified such that 40\% of his mass is in the head. Skeleton Zombie is much more massive at 146 kg and has 36\% of its mass in its arms alone (due to the giant claws). Our physics-based optimization can handle these variations and still accurately recover the motion from the video. Please see the supplementary video for additional examples.

%% file: appendix/tables/contact_f1.tex
\stepcounter{appendixtable}
\setlength{\tabcolsep}{4pt}
\begin{table}[t]
\caption{Precision, recall, and F1 Score (\textit{Prec/Rec/F1}) of estimating foot contacts from video. Left: comparison to various baselines, Right: ablations using subsets of joints as features. Supplements Table~\ref{table:contactestimation}.}
\vspace{-15pt}
\begin{center}
\scalebox{0.675}{
\begin{tabular}{ r c c | r c c }
\toprule
 \textbf{Baseline} & \textbf{Synthetic} & \textbf{Real} & \multicolumn{1}{c}{\textbf{MLP}} & \textbf{Synthetic} & \textbf{Real} \\
 \multicolumn{1}{r}{\textbf{Method}} & \textbf{Prec / Rec / F1} &  \textbf{Prec / Rec / F1} & \multicolumn{1}{c}{\textbf{Input Joints}} & \textbf{Prec / Rec / F1} & \textbf{Prec / Rec / F1} \\
\midrule
Random & 0.679 / 0.516 / 0.586 & 0.627 / 0.487 / 0.548 & Upper to hips & 0.940 / 0.941 / 0.940 & 0.728 / 0.837 / 0.779 \\ 
Always Contact & 0.677 / 1.000 / 0.808 & 0.647 / 1.000 / 0.786 & Upper to knees & \textbf{0.958} / 0.946 / 0.952 & 0.926 / 0.859 / 0.892 \\
2D Velocity & 0.861 / 0.933 / 0.896 & 0.922 / 0.868 / 0.894 & Lower to ankles & 0.933 / 0.971 / 0.952 & \textbf{0.963} / 0.916 / 0.939 \\ 
3D Velocity & 0.858 / 0.876 / 0.867 & 0.920 / 0.884 / 0.902 & Lower to hips & 0.941 / \textbf{0.973} / \textbf{0.957} & 0.956 / \textbf{0.943} / \textbf{0.949} \\
\bottomrule
\end{tabular}}
\end{center}
\vspace{-3pt}
\label{table:contactf1}
\end{table}

%% file: appendix/tables/contact_window.tex
\stepcounter{appendixtable}
\begin{wraptable}{r}{0.4\textwidth}
\vspace{-65pt}
\caption{Ablation study of input and output window sizes for learned contact estimation. Classification accuracy for many different combinations are shown.
}
\vspace{-15pt}
\begin{center}
\scalebox{0.65}{
\begin{tabular}{ c c c c }
\toprule
 Input & Prediction & Synthetic & Real \\
 \textbf{Window} & \textbf{Window} & \textbf{Accuracy} &  \textbf{Accuracy} \\
\midrule
3 & 3 & 0.931 & 0.919 \\
\midrule
5 & 3 & 0.933 & 0.913 \\
5 & 5 & 0.943 & 0.906 \\
\midrule
7 & 3 & 0.936 & 0.922 \\
7 & 5 & 0.941 & 0.923 \\
7 & 7 & 0.943 & 0.926 \\
\midrule
9 & 3 & 0.936 & 0.905 \\
9 & 5 & 0.941 & \textbf{0.935} \\
9 & 7 & 0.942 & 0.921 \\
9 & 9 & \textbf{0.946} & 0.927 \\
\bottomrule
\end{tabular}}
\end{center}
\vspace{-30pt}
\label{table:windowablation}
\end{wraptable}

%% file: appendix/figures/offtarget.tex
\stepcounter{appendixfigure}
\begin{figure}[t]
\begin{center}
\includegraphics[width=0.65\linewidth]{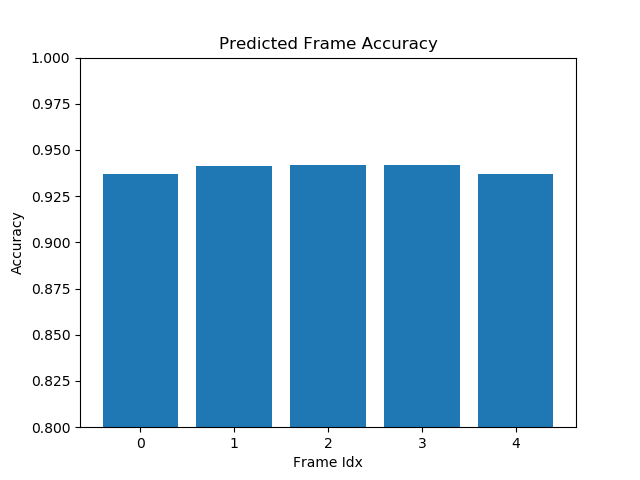}
\end{center}
\vspace{-15pt}
   \caption{Contact estimation classification accuracy for all frames in the 5-frame output window on the synthetic test set (given 9 frames as input). The center frame index 2 is the target frame, however off-target contact predictions are still accurate.
   }
\vspace{-15pt}
\label{fig:offtarget}
\end{figure}

%% file: appendix/tables/fitted_perception.tex
\stepcounter{appendixtable}
\begin{table*}[t]
\caption{Physical plausibility evaluation using the \emph{estimated floor} on the synthetic test set. Supplements Table~\ref{table:perceptioneval}.}
\vspace{-15pt}
\begin{center}
\scalebox{0.85}{
\begin{tabular}{ r | r r r | r r r }
\toprule
  & \multicolumn{3}{c}{Dynamics (Contact forces)} & \multicolumn{3}{c}{Kinematics (Foot positions)} \\
\midrule
 \textbf{Method} & \textbf{Mean GRF} & \textbf{Max GRF} & \textbf{Ballistic GRF} &  \textbf{Floating} & \textbf{Penetration} & \textbf{Skate} \\
\midrule
MTC \cite{mtc} & 142.7\% & 9036.7\% & 120.8\% & 19.1\%	& 10.0\% & 16.5\% \\
Kinematics (ours) & 119.7\% & 1252.4\% & 103.6\% & \textbf{1.5\%} & 1.8\% & \textbf{1.3\%} \\
Physics (ours) & \textbf{ 98.8\%} & \textbf{293.2\%} & \textbf{0.0\%} & 5.9\% & \textbf{0.1\%} & 3.8\% \\
\bottomrule
\end{tabular}}
\end{center}
\label{table:fittedperceptioneval}
\vspace{-10pt}
\end{table*}

%% file: appendix/tables/fitted_pose.tex
\stepcounter{appendixtable}
\setlength{\tabcolsep}{4pt}
\begin{wraptable}{r}{0.5\textwidth}
\vspace{-15pt}
\caption{Pose evaluation on synthetic test set using the \emph{estimated floor}. Supplements Table~\ref{table:poseestimation}.}
\vspace{-15pt}
\begin{center}
\scalebox{0.70}{
\begin{tabular}{ r | c c c }
\toprule
 \textbf{Method} & \textbf{Feet}  & \textbf{Body} &  \textbf{Body-Align 1} \\ 
\midrule
MTC \cite{mtc} & 585.303 & \textbf{565.068}	& \textbf{277.296} \\ 
Kinematics (ours) 	& 582.400 & 565.097 & 281.416 \\
Physics (ours) & \textbf{582.311} & 587.627 & 319.517 \\
\bottomrule
\end{tabular}}
\end{center}
\vspace{-5pt}
\label{table:fittedposeestimation}
\vspace{-20pt}
\end{wraptable}

%% file: appendix/figures/mtc_compare.tex
\stepcounter{appendixfigure}
\begin{figure}[t]
\begin{center}
\includegraphics[width=0.8\linewidth]{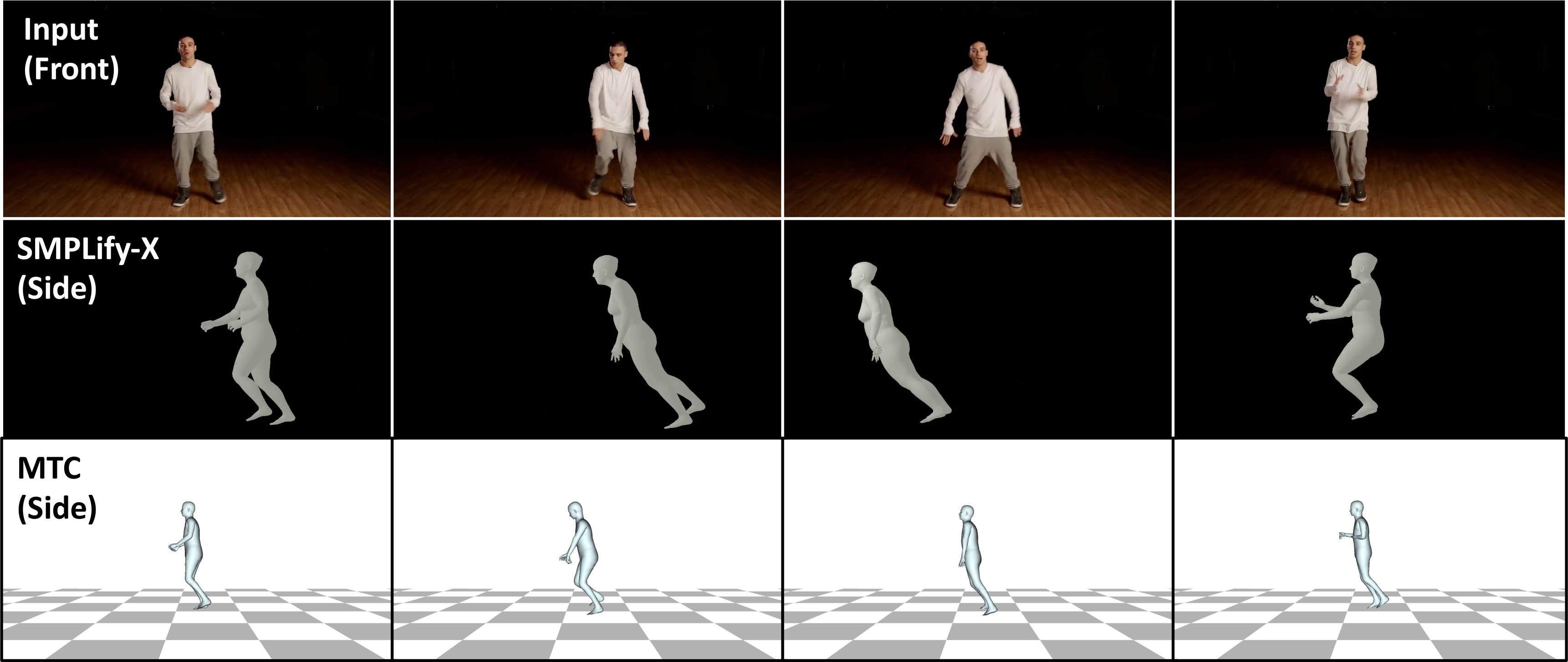}
\end{center}
\vspace{-15pt}
   \caption{A fixed side view is shown from SMPLify-X~\cite{SMPL-X:2019} and Monocular Total Capture (MTC)~\cite{mtc}. SMPLify-X gives noisy and inconsistent global motion whereas the tracking refinement of MTC gives smoother results.
   }
\label{fig:mtcsmplifycompare}
\end{figure}

%% file: appendix/figures/multi_char.tex
\stepcounter{appendixfigure}
\begin{figure}[t]
\begin{center}
\includegraphics[width=0.9\linewidth]{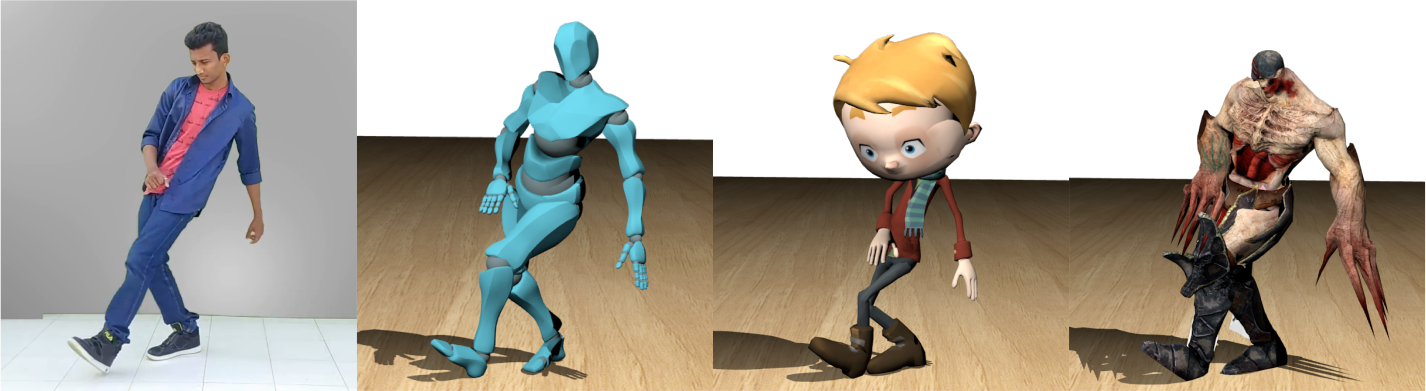}
\end{center}
\vspace{-15pt}
   \caption{Our method can be applied to multiple characters with varying body masses and mass distributions. From left to right the animated characters are Ybot (body mass 73 kg), Ty (36.5 kg), and Skeleton Zombie (146 kg).
   }
\label{fig:chars}
\end{figure}

%% file: eccv2020submissionCR.bbl
\begin{thebibliography}{10}
\providecommand{\url}[1]{\texttt{#1}}
\providecommand{\urlprefix}{URL }
\providecommand{\doi}[1]{https://doi.org/#1}

\bibitem{Bogo:ECCV:2016}
Bogo, F., Kanazawa, A., Lassner, C., Gehler, P., Romero, J., Black, M.J.: Keep
  it smpl: Automatic estimation of 3d human pose and shape from a single image.
  In: European Conference on Computer Vision (ECCV). pp. 561--578 (2016)

\bibitem{BrubakerICCV2009}
{Brubaker}, M.A., {Sigal}, L., {Fleet}, D.J.: Estimating contact dynamics. In:
  The IEEE International Conference on Computer Vision (ICCV). pp. 2389--2396
  (2009)

\bibitem{BrubakerIJCV2010}
Brubaker, M.A., Fleet, D.J., Hertzmann, A.: Physics-based person tracking using
  the anthropomorphic walker. International Journal of Computer Vision
  \textbf{87}(1),  140--155 (2010)

\bibitem{cao2018openpose}
Cao, Z., Hidalgo, G., Simon, T., Wei, S.E., Sheikh, Y.: Openpose: Realtime
  multi-person 2d pose estimation using part affinity fields. IEEE Transactions
  on Pattern Analysis and Machine Intelligence  (2019)

\bibitem{carpentier2018multicontact}
Carpentier, J., Mansard, N.: Multicontact locomotion of legged robots. IEEE
  Transactions on Robotics  \textbf{34}(6),  1441--1460 (2018)

\bibitem{chan2019dance}
Chan, C., Ginosar, S., Zhou, T., Efros, A.A.: Everybody dance now. In: IEEE
  International Conference on Computer Vision (ICCV) (2019)

\bibitem{chen2019holistic++}
Chen, Y., Huang, S., Yuan, T., Qi, S., Zhu, Y., Zhu, S.C.: Holistic++ scene
  understanding: Single-view 3d holistic scene parsing and human pose
  estimation with human-object interaction and physical commonsense. In: The
  IEEE International Conference on Computer Vision (ICCV). pp. 8648--8657
  (2019)

\bibitem{Choi:1999:OMR}
Choi, K.J., Ko, H.S.: On-line motion retargetting. In: Pacific Conference on
  Computer Graphics and Applications. pp.~32-- (1999)

\bibitem{dai2014whole}
Dai, H., Valenzuela, A., Tedrake, R.: Whole-body motion planning with
  centroidal dynamics and full kinematics. In: IEEE-RAS International
  Conference on Humanoid Robots. pp. 295--302 (2014)

\bibitem{Fang:2003:ESP}
Fang, A.C., Pollard, N.S.: Efficient synthesis of physically valid human
  motion. ACM Trans. Graph.  \textbf{22}(3),  417--426 (2003)

\bibitem{ForsythSurvey}
Forsyth, D.A., Arikan, O., Ikemoto, L., O'Brien, J., Ramanan, D.: Computational
  studies of human motion: Part 1, tracking and motion synthesis. Foundations
  and Trends in Computer Graphics and Vision  \textbf{1}(2–3),  77--254
  (2006)

\bibitem{Gleicher:1998:RMN}
Gleicher, M.: Retargetting motion to new characters. In: SIGGRAPH. pp. 33--42
  (1998)

\bibitem{hassan2019resolving}
Hassan, M., Choutas, V., Tzionas, D., Black, M.J.: Resolving 3d human pose
  ambiguities with 3d scene constraints. In: The IEEE International Conference
  on Computer Vision (ICCV). pp. 2282--2292 (2019)

\bibitem{he2017maskrcnn}
He, K., Gkioxari, G., Dollar, P., Girshick, R.: Mask r-cnn. In: The IEEE
  International Conference on Computer Vision (ICCV). pp. 2961--2969 (2017)

\bibitem{HerrPopovic}
Herr, H., Popovic, M.: Angular momentum in human walking. Journal of
  Experimental Biology  \textbf{211}(4),  467--481 (2008)

\bibitem{Hoyet:2012:PRP}
Hoyet, L., McDonnell, R., O'Sullivan, C.: Push it real: Perceiving causality in
  virtual interactions. ACM Trans. Graph.  \textbf{31}(4),  90:1--90:9 (2012)

\bibitem{Ikemoto:2006:KPY}
Ikemoto, L., Arikan, O., Forsyth, D.: Knowing when to put your foot down. In:
  Symposium on Interactive 3D Graphics and Games (I3D). pp. 49--53 (2006)

\bibitem{h36m_pami}
Ionescu, C., Papava, D., Olaru, V., Sminchisescu, C.: Human3.6m: Large scale
  datasets and predictive methods for 3d human sensing in natural environments.
  IEEE Transactions on Pattern Analysis and Machine Intelligence
  \textbf{36}(7),  1325--1339 (2014)

\bibitem{Jiang:2019:SBR}
Jiang, Y., Van~Wouwe, T., De~Groote, F., Liu, C.K.: Synthesis of biologically
  realistic human motion using joint torque actuation. ACM Trans. Graph.
  \textbf{38}(4),  72:1--72:12 (2019)

\bibitem{hmrKanazawa17}
Kanazawa, A., Black, M.J., Jacobs, D.W., Malik, J.: End-to-end recovery of
  human shape and pose. In: The IEEE Conference on Computer Vision and Pattern
  Recognition (CVPR). pp. 7122--7131 (2018)

\bibitem{humanMotionKanazawa19}
Kanazawa, A., Zhang, J.Y., Felsen, P., Malik, J.: Learning 3d human dynamics
  from video. In: The IEEE Conference on Computer Vision and Pattern
  Recognition (CVPR). pp. 5614--5623 (2019)

\bibitem{adam}
Kingma, D.P., Ba, J.: Adam: A method for stochastic optimization. In:
  International Conference on Learning Representations (ICLR) (2015)

\bibitem{KuligGRF}
Kulig, K., Fietzer, A.L., Jr., J.M.P.: Ground reaction forces and knee
  mechanics in the weight acceptance phase of a dance leap take-off and
  landing. Journal of Sports Sciences  \textbf{29}(2),  125--131 (2011)

\bibitem{deLasa:2010:FLC}
de~Lasa, M., Mordatch, I., Hertzmann, A.: Feature-based locomotion controllers.
  In: SIGGRAPH. pp. 131:1--131:10 (2010)

\bibitem{LeCallennec:2006:RKC}
Le~Callennec, B., Boulic, R.: Robust kinematic constraint detection for motion
  data. In: ACM SIGGRAPH/Eurographics Symposium on Computer Animation (SCA).
  pp. 281--290 (2006)

\bibitem{DELEVA19961223}
de~Leva, P.: Adjustments to zatsiorsky-seluyanov's segment inertia parameters.
  Journal of Biomechanics  \textbf{29}(9),  1223 -- 1230 (1996)

\bibitem{Li_2019_CVPR}
Li, Z., Sedlar, J., Carpentier, J., Laptev, I., Mansard, N., Sivic, J.:
  Estimating 3d motion and forces of person-object interactions from monocular
  video. In: The IEEE Conference on Computer Vision and Pattern Recognition
  (CVPR). pp. 8640--8649 (2019)

\bibitem{Liu:2005:LPM}
Liu, C.K., Hertzmann, A., Popovi\'{c}, Z.: Learning physics-based motion style
  with nonlinear inverse optimization. ACM Trans. Graph.  \textbf{24}(3),
  1071--1081 (2005)

\bibitem{Macchietto:2009:MCB}
Macchietto, A., Zordan, V., Shelton, C.R.: Momentum control for balance. In:
  SIGGRAPH. pp. 80:1--80:8 (2009)

\bibitem{VNect_SIGGRAPH2017}
Mehta, D., Sridhar, S., Sotnychenko, O., Rhodin, H., Shafiei, M., Seidel, H.P.,
  Xu, W., Casas, D., Theobalt, C.: Vnect: Real-time 3d human pose estimation
  with a single rgb camera. ACM Trans. Graph.  \textbf{36}(4) (2017)

\bibitem{newell2016stacked}
Newell, A., Yang, K., Deng, J.: Stacked hourglass networks for human pose
  estimation. In: European Conference on Computer Vision (ECCV). pp. 483--499
  (2016)

\bibitem{orin2013centroidal}
Orin, D.E., Goswami, A., Lee, S.H.: Centroidal dynamics of a humanoid robot.
  Autonomous Robots  \textbf{35}(2-3),  161--176 (2013)

\bibitem{pytorch}
Paszke, A., Gross, S., Massa, F., Lerer, A., Bradbury, J., Chanan, G., Killeen,
  T., Lin, Z., Gimelshein, N., Antiga, L., Desmaison, A., Kopf, A., Yang, E.,
  DeVito, Z., Raison, M., Tejani, A., Chilamkurthy, S., Steiner, B., Fang, L.,
  Bai, J., Chintala, S.: Pytorch: An imperative style, high-performance deep
  learning library. In: Advances in Neural Information Processing Systems
  (NeurIPS). pp. 8026--8037 (2019)

\bibitem{SMPL-X:2019}
Pavlakos, G., Choutas, V., Ghorbani, N., Bolkart, T., Osman, A.A.A., Tzionas,
  D., Black, M.J.: Expressive body capture: 3d hands, face, and body from a
  single image. In: The IEEE Conference on Computer Vision and Pattern
  Recognition (CVPR). pp. 10975--10985 (2019)

\bibitem{Pavllo:2019:posemetric}
Pavllo, D., Feichtenhofer, C., Grangier, D., Auli, M.: 3d human pose estimation
  in video with temporal convolutions and semi-supervised training. In: The
  IEEE Conference on Computer Vision and Pattern Recognition (CVPR). pp.
  7753--7762 (2019)

\bibitem{Peng:2018:SRL}
Peng, X.B., Kanazawa, A., Malik, J., Abbeel, P., Levine, S.: Sfv: Reinforcement
  learning of physical skills from videos. ACM Trans. Graph.  \textbf{37}(6),
  178:1--178:14 (2018)

\bibitem{Popovic:1999:PBM}
Popovi\'{c}, Z., Witkin, A.: Physically based motion transformation. In:
  SIGGRAPH. pp. 11--20 (1999)

\bibitem{Reitsma:2003:PMC}
Reitsma, P.S.A., Pollard, N.S.: Perceptual metrics for character animation:
  Sensitivity to errors in ballistic motion. ACM Trans. Graph.  \textbf{22}(3),
   537--542 (2003)

\bibitem{RMB}
Robertson, D.G.E., Caldwell, G.E., Hamill, J., Kamen, G., Whittlesey, S.N.:
  Research Methods in Biomechanics. Human Kinetics (2004)

\bibitem{Safonova:2004}
Safonova, A., Hodgins, J.K., Pollard, N.S.: Synthesizing physically realistic
  human motion in low-dimensional, behavior-specific spaces. In: SIGGRAPH. pp.
  514--521. ACM (2004)

\bibitem{Sigal:2010:HumanEva}
Sigal, L., Balan, A., Black, M.: Humaneva: Synchronized video and motion
  capture dataset and baseline algorithm for evaluation of articulated human
  motion. International Journal of Computer Vision  \textbf{87},  4--27 (2010)

\bibitem{Vondrak:2012:VMC}
Vondrak, M., Sigal, L., Hodgins, J., Jenkins, O.: Video-based 3d motion capture
  through biped control. ACM Trans. Graph.  \textbf{31}(4),  27:1--27:12 (2012)

\bibitem{ipopt}
W\"{a}chter, A., Biegler, L.T.: On the implementation of an interior-point
  filter line-search algorithm for large-scale nonlinear programming.
  Mathematical Programming  \textbf{106}(1),  25--57 (2006)

\bibitem{biolocomotion}
Wang, J.M., Hamner, S.R., Delp, S.L., Koltun, V.: Optimizing locomotion
  controllers using biologically-based actuators and objectives. ACM Trans.
  Graph.  \textbf{31}(4) (2012)

\bibitem{Wei:2010:VMP}
Wei, X., Chai, J.: Videomocap: Modeling physically realistic human motion from
  monocular video sequences. In: SIGGRAPH. pp. 42:1--42:10 (2010)

\bibitem{winkler18}
Winkler, A.W., Bellicoso, D.C., Hutter, M., Buchli, J.: Gait and trajectory
  optimization for legged systems through phase-based end-effector
  parameterization. IEEE Robotics and Automation Letters (RA-L)  \textbf{3},
  1560--1567 (2018)

\bibitem{mtc}
Xiang, D., Joo, H., Sheikh, Y.: Monocular total capture: Posing face, body, and
  hands in the wild. In: The IEEE Conference on Computer Vision and Pattern
  Recognition (CVPR). pp. 10965--10974 (2019)

\bibitem{Xu:2018:MHP}
Xu, W., Chatterjee, A., Zollh\"{o}fer, M., Rhodin, H., Mehta, D., Seidel, H.P.,
  Theobalt, C.: Monoperfcap: Human performance capture from monocular video.
  ACM Trans. Graph.  \textbf{37}(2),  27:1--27:15 (2018)

\bibitem{zanfir2018monocular}
Zanfir, A., Marinoiu, E., Sminchisescu, C.: Monocular 3d pose and shape
  estimation of multiple people in natural scenes-the importance of multiple
  scene constraints. In: The IEEE Conference on Computer Vision and Pattern
  Recognition (CVPR). pp. 2148--2157 (2018)

\bibitem{Zou_2020_WACV}
Zou, Y., Yang, J., Ceylan, D., Zhang, J., Perazzi, F., Huang, J.B.: Reducing
  footskate in human motion reconstruction with ground contact constraints. In:
  The IEEE Winter Conference on Applications of Computer Vision (WACV). pp.
  459--468 (2020)

\end{thebibliography}
